\documentclass[journal]{IEEEtran}
\usepackage{amsmath,amssymb,amsfonts,bm,mathtools,amsthm}
\usepackage{color,xcolor,url,cite,hyperref,cleveref}
\usepackage{booktabs}
\usepackage{tabularx}
\usepackage{algorithm,algpseudocode,listings}
\usepackage{array}
\usepackage[caption=false,font=normalsize,labelfont=sf,textfont=sf]{subfig}
\usepackage{multirow,threeparttable} 
\usepackage{cleveref}       
\usepackage{textcomp}
\usepackage{stfloats}
\usepackage{verbatim}
\usepackage{graphicx}

\DeclareMathOperator*{\argmin}{arg\,min}
\newtheorem{assumption}{Assumption}

\newtheorem{proposition}{Proposition}
\newtheorem{corollary}{Corollary}
\newtheorem{theorem}{Theorem}
\newtheorem*{remark}{Remark}

\crefname{equation}{}{}
\Crefname{equation}{Equation}{Equations}
\crefname{assumption}{assumption}{assumptions}
\Crefname{assumption}{Assumption}{Assumptions}
\crefname{lstlisting}{listing}{listings}
\Crefname{lstlisting}{Listing}{Listings}

\begin{document}

\title{Alada: Alternating Adaptation of Momentum Method for Memory-Efficient Matrix Optimization}

\author{Xiaoyu He, Yu Cai, Jin Jia, Canxi Huang, Wenqing Chen, Zibin Zheng
\thanks{All authors are with School of Software Engineering, Sun Yat-sen University, Zhuhai 519082, P. R. China. }
\thanks{E-mail: hexy73@mail.sysu.edu.cn (X. He), caiy86@mail2.sysu.edu.cn (Y. Cai), jiaj9@mail2.sysu.edu.cn (J. Jin), huangcx23@mail2.sysu.edu.cn (C. Huang), chenwq95@mail.sysu.edu.cn (W. Chen), zhzibin@mail.sysu.edu.cn (Z. Zheng)} 
\thanks{* Corresponding Author: W. Chen}
}
\markboth{Journal of \LaTeX\ Class Files,~Vol.~14, No.~8, August~2021}%
{He \MakeLowercase{\textit{et al.}}: Alada: Alternating Adaptation of Momentum Method for Memory-Efficient Matrix Optimization}


\maketitle

\begin{abstract}
  This work proposes Alada, an adaptive momentum method for stochastic optimization over large-scale matrices. 
	Alada employs a rank-one factorization approach to estimate the second moment of gradients, where factors are updated alternatively to minimize the estimation error.
	Alada achieves sublinear memory overheads and can be readily extended to optimizing tensor-shaped variables.
	We also equip Alada with a first moment estimation rule, which enhances the algorithm's robustness without incurring additional memory overheads.
	The theoretical performance of Alada aligns with that of traditional methods such as Adam.
	Numerical studies conducted on several natural language processing tasks demonstrate the reduction in memory overheads and the robustness in training large models relative to Adam and its variants.
\end{abstract}

\begin{IEEEkeywords}
Memory-efficient optimization, adaptive optimization methods, matrix optimization
\end{IEEEkeywords}

\section{Introduction}
We consider the following stochastic optimization problem over matrices
\begin{equation}\label{eq:definition}
	\min \limits_{X\in \mathbb{R}^{m\times n}} f(X) = \mathbb{E}_{\xi \sim \mathcal{D}}[F(X;\xi)]
\end{equation}
where $\xi$ denotes a data sample, $\mathcal{D}$ is the data distribution, and $X$ is the matrix-shaped decision variable.
Many machine learning tasks appear naturally in this form.
For example, in an $m$-class softmax regression problem, the objective is given by 
\[
F(X;\xi) = \ell_\text{CE}(Xy,z)
\]
where $\ell_\text{CE}$ is the cross-entropy loss, and the data sample $\xi$ consists of an $n$-dimensional feature vector $y$ and an $m$-dimensional one-hot vector $z$ denoting its target label.
Another example is the multilayered neural network~\cite{higham_deep_2019} which, on receiving an $n_0$-dimensional vector $y$, computes
\[
\sigma(X^{[L]} \dots\sigma(X^{[2]} \sigma(X^{[1]} y)))
\]
where $X^{[l]} \in \mathbb{R}^{n_l\times n_{l-1}}$ (with $n_0=n$) is the weight matrix of the $l$-th fully connected layer, $\sigma$ denotes some activation function applied element-wisely, and $L$ is the number of layers. 
In a typical training loop, these matrices are updated one-by-one, and each update on $X^{[l]}$ is equivalent to solving a subproblem~\cite{loizou_momentum_2020} over $n_l \times n_{l-1}$ matrices that can be written in the form of \cref{eq:definition}.

Adaptive methods such as Adam~\cite{kingma_adam:_2015} and AdaGrad~\cite{duchi_adaptive_2011} are the workhorse for solving problem \cref{eq:definition}.
These methods perform the update at iteration $t$ as
\begin{equation}\label{eq:adam-descent-update}
	X_{t+1} = X_t - \eta_t \frac{M_{t+1}}{\sqrt{U_{t+1} + \epsilon}}
\end{equation}
where $\eta_t \in \mathbb{R}_+$ is the step-size, $\epsilon \in \mathbb{R}_+$ is a small constant,
and $M_{t+1}$ and $U_{t+1}$ are momenta estimating the first and second moments of gradients respectively.
The term $M_{t+1}$ is usually intended to improve the robustness against data noise while the use of $U_{t+1}$ helps in handling landscape ill-conditioning.
Note that the division and the square root operators are performed element-wisely, so \cref{eq:adam-descent-update} is essentially a diagonal-approximation-based second-order descent step.
The element-wise pre-conditioning makes the algorithm easy to implement and applicable to different shapes of decision variables. 
However, since the momenta $M_{t+1}$ and $U_{t+1}$ need to be maintained explicitly over iterations, extra overheads\footnotemark of $2mn$ memory usage are introduced.
This limits the applicability of adaptive methods in large-scale settings.
\footnotetext{We denote the memory overhead of an adaptive method by the additional memory usage that is 1) not required by the standard stochastic gradient method~\cite{ghadimi_stochastic_2013}, and 2) for storing optimizer states that must be maintained over iterations. For example, the overhead of Adam is $2mn$ because it stores the terms $M_{t+1}$ and $U_{t+1}$. Temporary variables such as $(U_{t+1} + \epsilon)$ in \cref{eq:adam-descent-update} are not counted because their memory can be freed immediately after each update.}

We describe a new  method called \underline{al}ternating \underline{ada}ptation (Alada) to address the above challenge.  
Alada retains competitive performance with state-of-the-art methods such as Adam, while reducing their memory overheads from $\mathcal{O}(mn)$ to $\mathcal{O}(m+n)$.
It is based on recent studies on rank-one momentum factorization~\cite{shazeer_adafactor_2018,luo_came_2023}, where the second moment of gradients is estimated by the outer product of two vector-shaped momenta. 
We propose an alternating minimization based method to adapt these vectors, ensuring that the estimation error can reduce in each iteration.  
We further show that Alada can estimate the first moment with no extra cost in memory. To the best of our knowledge, this is the first adaptive algorithm achieving sublinear memory overheads ($\mathcal{O}(m+n)$) while implementing both first and second moment estimations.
Theoretical analysis shows that Alada achieves similar convergence properties compared to Adam. 
Experiments on several natural language processing tasks verify the effectiveness of Alada. 

\paragraph*{Notation}
$\|\cdot\|$ denotes the 2-norm of vectors and F-norm of matrices.
$\|\cdot\|_\infty$ denotes the infinity norm of matrices, i.e., $\|A\|_\infty = \max\limits_{i,j}\{|A_{i,j}|\}$.
$1_m$ and $1_{m\times n}$ denote the all-one vector of length $m$ and the all-one matrix of size $m\times n$, respectively.
$0_m$ and $0_{m\times n}$ denote the all-zero vector of length $m$ and the all-zero matrix of size $m\times n$, respectively.

\section{Preliminaries on adaptive methods}
In the Adam-family of adaptive methods, the first and second moments of gradients are estimated by momenta that are exponential moving averages in the form of  
\begin{equation}\label{eq:adam-adaptation-rule} 
\begin{cases}
	M_{t+1} = \beta_1 M_t + (1-\beta_1)G_t \\
	U_{t+1} = \beta_2 U_t + (1-\beta_2)G^2_t
\end{cases}
\end{equation}
where $G_t$ is some stochastic gradient estimate, and the square operator is performed element-wisely. 
We assume the decay parameters $\beta_1,\beta_2 \in [0,1)$ are fixed for simplicity.
In its simplest form, $G_t$ is a Monte Carlo estimate as 
\[
	G_t = \nabla F(X_t;\xi_t) \text{ with } \xi_t \sim \mathcal{D}.
\]
Therefore, the operation of dividing $M_{t+1}$ by the squared root of $U_{t+1}$ can be viewed as a kind of variance reduction procedure, which fixes the scaling issue caused by noise~\cite{kunstner_limitations_2019}.
This usually improves the convergence performance, and is perhaps the main reason behind the popularity of adaptive algorithms.
The disadvantage of this design is however obvious: the algorithm needs to maintain two momenta, yielding $\mathcal{O}(mn)$ memory usage.

The Adafactor method~\cite{shazeer_adafactor_2018} is one of the pioneering works that address the memory issue. 
Its key idea is to factorize the second moment estimate as $G_t^2 \approx p q^\top$ and then accumulate the vectors $p$ and $q$ individually.
These vectors are found through minimizing the KL-divergence that measures the factorization error, leading to the following adaptation rule:
\begin{equation}\label{eq:adafactor-update-rule}
\begin{cases}
	(p_t^*,q_t^*) \in \argmin \limits_{p,q} \text{KL}(G_t^2 \parallel p q^\top), \\
	(p_{t+1},q_{t+1}) = \text{update }(p_t,q_t) \text{ with } (p_t^*,q_t^*).
\end{cases}
\end{equation}
This method avoids explicitly updating the momentum $U_{t+1}$ as in \cref{eq:adam-adaptation-rule}, thereby saving $mn$ memory usage.
The authors of \cite{shazeer_adafactor_2018} further suggested setting $\beta_1 = 0$ in \cref{eq:adam-adaptation-rule}. This means the stochastic gradient $G_t$ rather than the momentum $M_{t+1}$ is used for guiding the search, which saves additional $mn$ memory usage.
Therefore, the descent step of Adafactor reads 
\[
	X_{t+1} = X_t - \eta_t \frac{G_t}{\sqrt{\text{rec}(p_{t+1},q_{t+1}) + \epsilon}}
\]
where ``rec'' denotes recovering a matrix-shaped momentum term from vectors $p_{t+1}$ and $q_{t+1}$.
Due to its memory efficient nature, Adafactor has been widely used in training or fine-tuning language models where memory usage is the core bottleneck.

\section{Alternating adaptation of the second moment}
\label{sec:alternating-adaptation-of-2nd-moment}
We describe a new method to adapt the second moment of gradients with sublinear memory usage.
The method is similar to Adafactor in that it is based on the factorization $G_t^2 \approx p q^\top$, but differs in how the factorization and adaptation are performed.
Specifically, we propose to perform factorization via minimizing the Euclidean distance: 
\begin{equation}\label{eq:factorization-via-minimizing-euclidean-distance}
	\argmin \limits_{p,q} \left\|G_t^2 - p q^\top\right\|^2.
\end{equation}
The purpose of using the Euclidean distance rather than the KL-divergence is twofold.
First, the terms $G_t^2$ and $pq^\top$ are not necessarily in the simplex, so the meaning of computing the KL-divergence between them might be ambiguous.
Second, the Euclidean distance is symmetric, which is not the case of KL-divergence.
Therefore, the Euclidean distance offers a more reasonable metric in measuring the factorization error.

A technical difficulty remains in solving the problem \cref{eq:factorization-via-minimizing-euclidean-distance}:
the objective function is not convex in $(p,q)$ and solving the problem exactly is known to require the singular value decomposition of $G_t^2$.
Fortunately, we can minimize \cref{eq:factorization-via-minimizing-euclidean-distance} inexactly based on the fact that it is convex in $p$ and in  $q$ individually.
For example, we can fix $q$, solve for $p$, and then use the obtained solution to update $p$.
This results in the update rule:
\begin{equation}\label{eq:update-p}
\begin{cases}
	p_t^* = \argmin \limits_p \left\|G_t^2 - pq_t^\top\right\|^2, \\
	p_{t+1} = \text{update }p_t \text{ with } p_t^*.
\end{cases}
\end{equation}	
Alternatively, we can derive an update rule for $q$ via fixing $p$:
\begin{equation}\label{eq:update-q}
\begin{cases}
	q_t^* = \argmin \limits_q \left\|G_t^2 - p_tq^\top\right\|^2, \\
	q_{t+1} = \text{update }q_t \text{ with } q_t^*.
\end{cases}
\end{equation}	
When $p$ and $q$ are strictly positive, $\left\|G_t^2 - pq^\top\right\|^2$ is strongly convex in $p$ and in $q$, respectively. 
In this case, $p_t^*$ and $q_t^*$ are unique and are given by $p_t^* = G_t^2 q_t / \|q_t\|^2$ and $q_t^* = \left(G_t^2\right)^\top p_t / \|p_t\|^2$, respectively.
Our idea is to alternate between \cref{eq:update-p} and \cref{eq:update-q} over iterations to avoid solving \cref{eq:factorization-via-minimizing-euclidean-distance} exactly (which is time-consuming). 
In addition, when all moment estimates $\{G_t^2\}$ are positive, the iterations $\{p_t\}$ and $\{q_t\}$ would remain positive if their initial values are. 
This means the outer product $p_t q_t^\top$ is a feasible estimate to the second moment, so they serve as a preconditioner in performing descent steps.

The above idea is implemented in \Cref{alg:Alada-informal}.
While Adam involves a matrix-shaped momentum $U_t$, Alada maintains two vector-shaped momenta $p_t$ and $q_t$ instead, reducing the memory overhead to $\mathcal{O}(m+n)$.
In the main loop, Alada keeps $p_t$ ($q_t$) unchanged while updating $q_t$ ($p_t$) with $t$ odd (even). 
Here the update is controlled by a decay parameter $\beta_2 \in [0,1)$.
If letting $U_t = p_t q_t^\top$, we recover
\begin{equation}\label{eq:low-rank-variance-accumulation}
	U_{t+1} = 
	\begin{cases}
		\begin{aligned}
		\beta_2 U_t + (1-\beta_2) G_t^2\frac{q_t q_t^\top}{\|q_t\|^2},& \qquad t \text{ is even}, \\[1ex]
		\beta_2 U_t + (1-\beta_2) \frac{p_t p_t^\top}{\|p_t\|^2} G_t^2,& \qquad t \text{ is odd}.
		\end{aligned}
	\end{cases}
\end{equation}
Comparing this with \cref{eq:adam-adaptation-rule}, it is found that Alada is a low-rank variant of Adam where a projection of the gradient variance, rather than the original one, is accumulated. 
This reveals the connection between Alada and Adam.

On the other hand, the $t$-th update of Alada ensures decreasing the factorization error w.r.t. the gradient variance $G_t^2$. 
We formulate this property below.
This follows directly from the convexity of the subproblems in \cref{eq:update-p,eq:update-q}, so we omit the proof.
\begin{proposition}
	The iterations $\{U_t\}$ computed according to \cref{eq:low-rank-variance-accumulation} satisfy
	\[
		\left\|G_t^2 - U_{t+1}\right\| \le \left\|G_t^2 - U_t\right\|.
	\]
\end{proposition}

\begin{figure}[thb]
\begin{algorithm}[H]
	\begin{algorithmic}[1]
	\caption{Alada (informal version)}
	\label{alg:Alada-informal}
	\For {$t \leftarrow 0,1,2, \dots$} 
		\State Draw a sample $\xi_t$ from $\mathcal{D}$
		\State $G_t = \nabla F(X_t;\xi_t)$
		\If {$t\in\{0,2,4,\dots\}$}
			\State $p_{t+1} = \beta_2 p_t + (1-\beta_2) \frac{G_t^2 q_t}{\|q_t\|^2}$
			\State $q_{t+1} = q_t$
		\Else
			\State $p_{t+1} = p_t$
			\State $q_{t+1} = \beta_2 q_t + (1-\beta_2) \frac{(G_t^2)^\top p_t}{\|p_t\|^2}$
		\EndIf
		\State $X_{t+1} = X_t - \eta \frac{G_t}{\sqrt{p_{t+1} q_{t+1}^\top+\epsilon}}$
	\EndFor
\end{algorithmic} 
 \end{algorithm}
\end{figure}

\section{Implementation}
We detail the implementations of the Alada method.

\subsection{First moment estimation}
The alternating adaptation method discussed so far achieves sublinear memory overheads, yet still misses a first moment estimation mechanism.
To further improve the performance of Alada, we adopt the same rule as in Adam to estimate the first moment: 
\begin{equation}\label{eq:1st-order-momentum}
M_{t+1} = \beta_1 M_t + (1-\beta_1)G_t
\end{equation}
where $\beta_1 \in (0,1)$ is an additional decay parameter and the initial momentum is set to $M_0 = 0_{m\times n}$.
In addition, we correct the bias introduced by the initialization as 
\begin{equation}\label{eq:bias-correction-first-order-momentum}
\tilde{M}_{t+1} = M_{t+1} / (1-\beta_1^{t+1}).
\end{equation}
The above offers a robust gradient estimate which can be used in guiding the search.
This, however, introduces additional memory overheads: $mn$ memory is required for storing the momentum term $M_{t+1}$.\footnote{Alternatively, one may maintain the bias-corrected momentum $\tilde{M}_{t+1}$; but this still requires $mn$ memory usage.}

Here we show how to estimate the first moment in Alada without causing significant overheads. 
The idea is to modify the alternating adaptation mechanism in \Cref{sec:alternating-adaptation-of-2nd-moment} such that the second moment is estimated from the variance of first moment estimates instead of the original gradient.
This allows discarding the gradient immediately after it is used for updating the momentum term $M_{t+1}$, thereby saving $mn$ memory usage.
The pseudocode is given in \Cref{alg:Alada-formal}.
In lines 3-6, we first generate a stochastic gradient estimate, obtain momentum $M_{t+1}$ via the exponential moving average, and then apply the bias-correction rule. 
In line 7, we use the variance of the bias-corrected momentum term $\tilde{M}_{t+1}$ as an approximation of the gradient variance estimate, denoted by $V_{t+1}$. 
It is then clear that the gradient $G_t$ is not required in subsequent steps.

\begin{figure}[thb]
\begin{algorithm}[H]
	\caption{Alada (practical implementation)}
	\label{alg:Alada-formal}
	\begin{algorithmic}[1]
	\Require initial solution $X_0 \in \mathbb{R}^{m\times n}$; step-size sequence $\{\eta_t\}$; decay parameters $\beta_1,\beta_2 \in (0,1)$; constant $\epsilon=10^{-16}$
	\State $M_0 \gets 0_{m\times n}$
	\For {$t \leftarrow 0,1,2, \dots$} 
		\State Draw a sample $\xi_t$ from $\mathcal{D}$
		\State $G_t = \nabla F(X_t;\xi_t)$
		\State $M_{t+1} = \beta_1 M_t + (1-\beta_1) G_t$
		\State $\tilde{M}_{t+1} = M_{t+1}/(1-\beta_1^{t+1})$
		\State $V_{t+1} = \tilde{M}_{t+1}^2$
		\If {$t=0$}
			\State $v_0 = \|G_0\|^2 / mn$
			\State $p_0 = \sqrt{v_0}1_m$
			\State $q_0 = \sqrt{v_0}1_n$
		\EndIf
		\If {$t\in\{0,2,4,\dots\}$}
			\State $p_{t+1} = \beta_2 p_t + (1-\beta_2) \frac{V_{t+1} q_t}{\|q_t\|^2+\epsilon}$
			\State $q_{t+1} = q_t$
		\Else
			\State $p_{t+1} = p_t$
			\State $q_{t+1} = \beta_2 q_t + (1-\beta_2) \frac{V_{t+1}^\top p_t}{\|p_t\|^2+\epsilon}$
		\EndIf
		\State $U_{t+1} = p_{t+1} q_{t+1}^\top$
		\State $\tilde{U}_{t+1} = (U_{t+1} - \beta_2^{t+1} v_0 )/(1-\beta_2^{t+1})$
		\State $X_{t+1} = X_t - \eta_t \frac{\tilde{M}_{t+1}}{\sqrt{\tilde{U}_{t+1}+\epsilon}}$
	\EndFor
\end{algorithmic} 
\end{algorithm}
\end{figure}

Now we describe how the above modification admits reducing memory overhead.
It is known that most practical optimization tasks are performed on automatic differentiation (AD) platforms, so the gradient is usually stored in leaf nodes of a computation graph and can be obtained through backpropagation.
When implementing the Alada method, instead of generating the gradient and then freeing it after use, we can directly store the first moment estimate $\tilde{M}_t$ in the memory slots that are intended to hold the gradient $G_t$.
The gradient obtained from backpropagation can then be accumulated into the momentum, and hence, will not introduce additional memory cost.
Popular AD platforms such as PyTorch~\cite{paszke_pytorch_2019} have built-in support for implementing this idea.
\Cref{lst:torch-code-Alada} shows a PyTorch-styled code snippet.  
Suppose here that we are training some machine learning model denoted by \texttt{mdl}.
The call \texttt{mdl.parameters()} in line 3 returns the trainable weights that will be updated, and they are the leaf nodes of the computation graph built by the AD engine.
In PyTorch, every leaf node possesses a field called \texttt{grad} for storing its associated gradient. 
In line 5, the call \texttt{mdl.zero\_grad()} initializes all \texttt{grad}-fields to zero.
Note that this differs from typical training processes which reset these fields every time before obtaining new gradients.
Therefore, in the main training loop, we can first scale down the \texttt{grad}-field of every leaf node by a factor of $\beta_1$ (in line 10) and then compute the objective value \texttt{fval1} (in lines 12-14), which is returned from some loss function.
We further multiply \texttt{fval1} by $(1-\beta_1)$ and obtain a scaled objective value \texttt{fval2}.
By performing backpropagation from \texttt{fval2}, the gradient w.r.t. \texttt{fval1} is computed, then scaled by $(1-\beta_1)$, and finally added to the \texttt{grad}-field of all leaf nodes.
This ensures the values in the \texttt{grad}-field are exactly the momentum given in \cref{eq:1st-order-momentum}.
In this way, Alada can enjoy the benefits of using first moment estimation without introducing additional memory overheads.  

\lstset{
	language=Python,
	frame=single,
	caption={An example of computing the first moment estimate with PyTorch},
	basicstyle=\footnotesize\ttfamily,
	captionpos=b,
	numbers = left,
	breaklines=true, 
	keywordstyle=\bfseries\color{black},
	tabsize=4,
	label=lst:torch-code-Alada,
}
\begin{figure}[htb]
\begin{lstlisting}
mdl = some_model()							
# specify decision variables
solver = Alada(mdl.parameters())			
# set the initial momentum to 0
mdl.zero_grad(set_to_none = False)			
# draw data samples
for feature, target in some_dataloader:		
	for para in mdl.parameters():			
      # scale down the previous momentum
		para.grad *= beta1					
    # forward pass
	output = mdl(feature)					
    # compute the objective value
	fval1 = compute_loss(output, target) 	
    # scale down the objective value
	fval2 = fval1 * (1-beta1)				
    # accumulate the scaled gradient
	fval2.backward()						
    # perform subsequent steps, e.g., updating second moment and performing descent
	solver.step()							
\end{lstlisting}
\end{figure}

\begin{figure}[htb]
\centering
\subfloat[Adafactor and Adam]{\includegraphics[width=0.48\textwidth]{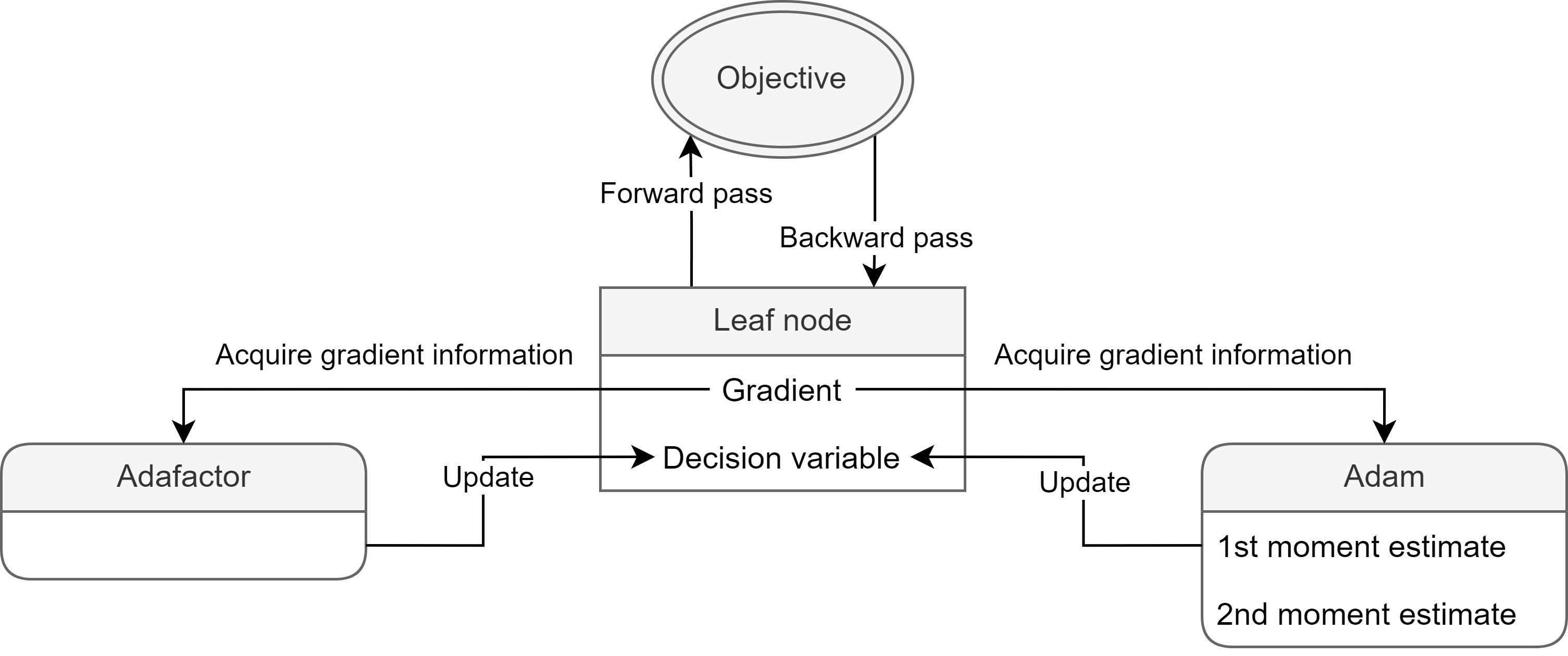}} \\
\subfloat[Alada]{\includegraphics[width=0.4\textwidth]{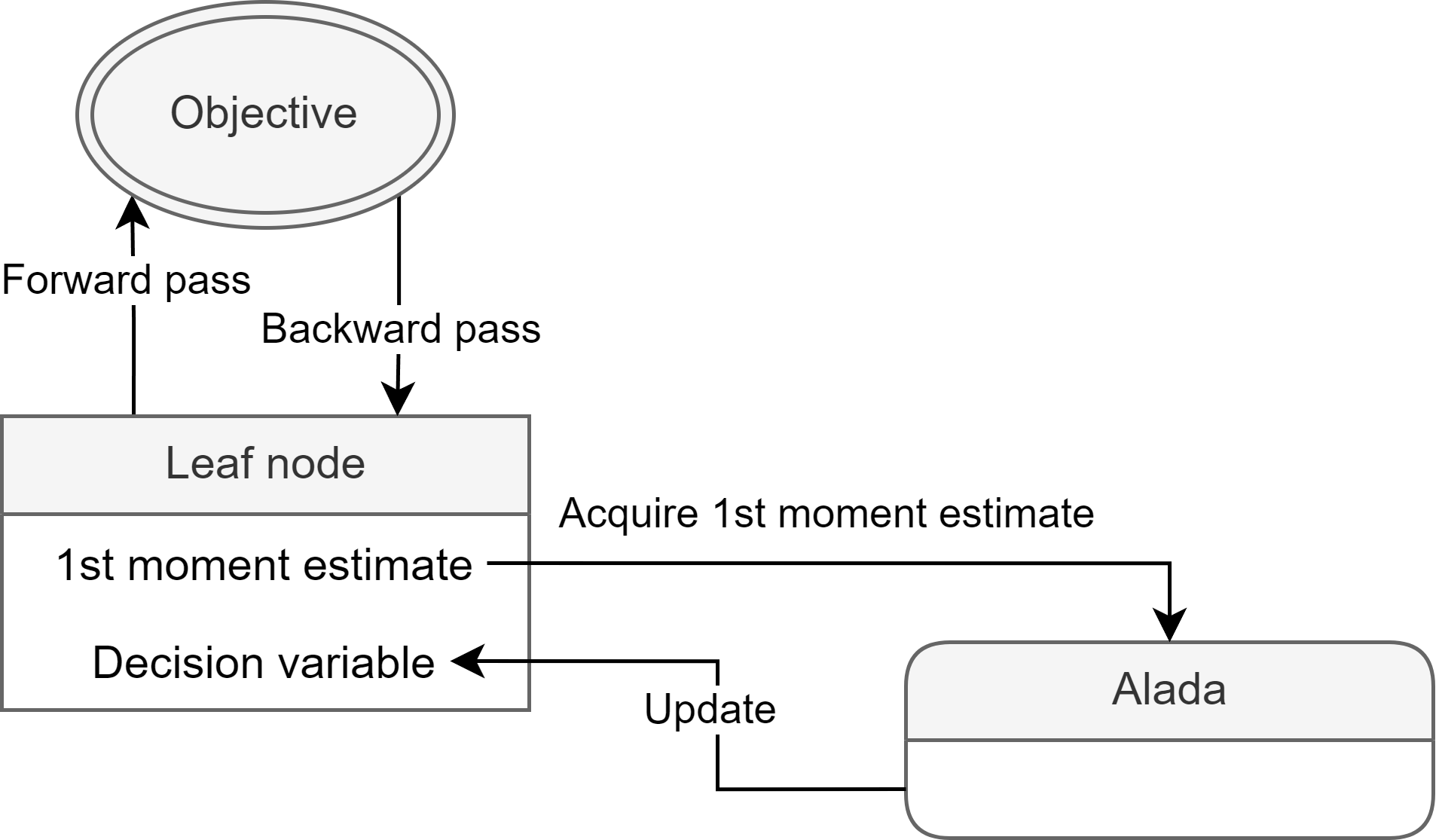}}
\caption{Memory layout when running Adafactor, Adam, or Alada with AD platforms like PyTorch.
The above only depicts variables that need to be maintained over iterations \protect\footnotemark ~and occupy at least $O(mn)$ memory slots. 
Adam maintains two momenta, each taking $mn$ memory slots.
Adafactor estimates the second moment by an outer product of two vectors and does not involve a first moment estimation mechanism. Therefore, its memory usage for maintaining momentum is negligible.
Alada is similar to Adafactor in terms of the peak memory usage. 
However, it admits estimating the first moment without significant memory overhead. 
It is because in Alada the gradient is not used directly when estimating the second moment or performing the descent step, so the gradient can be accumulated into the previous momentum stored in leaf nodes.
}
\label{fig:memory-layout}
\end{figure}

\Cref{fig:memory-layout} shows the difference among Alada, Adam, and Adafactor in terms of the memory layout.
In Adam, the momentum estimating the first moment is an optimizer internal state, while in Alada it is handled by the AD platform.
Adafactor discards this momentum term to reduce memory cost.
Alada implements the first moment estimation mechanism as Adam, but has the same memory requirements as Adafactor.

\footnotetext{The gradient associated with the leaf node when running Adafactor or Adam can be freed once not required. However, this does not reduce the peak memory usage, so we consider it as  maintained over iterations.}

\subsection{Initialization and bias correction}
\label{ss:initialization-bias-correction-lazy-preconditioning}
In Alada, the momentum estimating the first moment, $M_{t+1}$, is in the matrix form and is initialized with $M_0 = 0_{m\times n}$ as suggested by Adam.
The one estimating the second moment, $U_{t+1}$, is not maintained explicitly, but reconstructed from the vectors $p_{t+1}$ and $q_{t+1}$.
We initialize these vectors by setting $p_0 = \sqrt{v_0}1_m$ and $q_0 = \sqrt{v_0}1_n$, where $v_0 = \|G_0\|^2/mn$; see lines 8-12 of \Cref{alg:Alada-formal}.
This allows the initial momentum to have the same magnitude as the variance of the initial gradient. 

Now we verify the bias introduced due to the above initialization method.
Let $U_{t+1} = p_{t+1} q_{t+1}^\top$ be the reconstructed estimate to the second moment.
By the alternating adaptation rule, we have
\[
U_{t+1} = p_{t+1}q_{t+1}^\top =
\begin{cases}
\beta_2 U_t + (1-\beta_2) p_t^* q_t^\top,& \; t \text{ is even}, \\	
\beta_2 U_t + (1-\beta_2) p_t (q_t^*)^\top,& \; t \text{ is odd},
\end{cases}
\]
where $p_t^* = \argmin \limits_{p}\|pq_t - \tilde{M}_{t+1}^2\|^2$ and $q_t^* = \argmin \limits_{q}\|p_t q - \tilde{M}_{t+1}^2\|^2$.
Therefore, we have $p_t^*q_t^\top \approx \tilde{M}_{t+1}^2$ with $t$ even and $p_t (q_t^*)^\top \approx \tilde{M}_{t+1}^2$ with $t$ odd.
It follows that 
\begin{equation}\label{eq:91054}
  \begin{split}
U_{t+1} & \approx \beta_2 U_t + (1-\beta_2) \tilde{M}_{t+1}^2 \\ 
& = \beta_2^{t+1} U_0 + (1-\beta_2) \sum_{j=1}^{t+1} \beta_2^{t+1-j} \tilde{M}_j^2.
  \end{split}
\end{equation}
As $\tilde{M}_j$ is the bias-corrected momentum for estimating the first moment, $\tilde{M}_j^2$ is expected to be close to the true second moment. 
Suppose now $\tilde{M}_j^2$ is fixed in expectation, i.e., $\mathbb{E}[\tilde{M}_j^2] = M$ for all $j$ and some $M$.
We have
\[
\begin{split}
\mathbb{E}[U_{t+1}] &\approx \beta_2^{t+1} p_0 q_0^\top + (1-\beta_2) \sum_{j=1}^{t+1} \beta_2^{t+1-j} M \\
	& = \beta_2^{t+1} v_0 1_{m\times n} + (1-\beta_2^{t+1}) M \\
	& \Rightarrow M \approx \frac{\mathbb{E}[U_{t+1}]-\beta_2^{t+1}v_0}{1-\beta_2^{t+1}}.
\end{split}
\] 
This reveals how the momentum term $U_{t+1}$ is related to the true second moment $M$. 
Therefore, in the practical implementation of Alada, we correct the term $U_{t+1}$ as 
\[
\tilde{U}_{t+1} = \frac{U_{t+1}-\beta_2^{t+1}v_0}{1-\beta_2^{t+1}}.
\]
We use the bias-corrected momentum $\tilde{U}_{t+1}$ to precondition the bias-corrected momentum $\tilde{M}_{t+1}$ and obtain a preconditioned descent step $W_{t+1}$. The latter is then used as a descent step; see lines 20-22 of \Cref{alg:Alada-formal}.

\subsection{Decay parameter settings}
\label{ss:decay-parameter-settings}
We describe how to set the decay parameters in Alada.
Since Alada is a memory efficient variant of Adam, we recommend setting these parameters such that Alada and Adam could perform similarly.
On one hand, to mimic the behavior of an Adam instance with some $\beta_1 = \beta_1^\text{Adam}$, we can use the same setting in Alada, since they employ exactly the same rule in estimating the first moment. 
The case of choosing $\beta_2$, on the other hand, could be different.
Let us consider how the gradient variance $G_t^2$ impacts the second moment estimation in Alada.
Firstly, using \cref{eq:1st-order-momentum,eq:bias-correction-first-order-momentum} and the setting $M_0 = 0_{m\times n}$, we have
\[
\tilde{M}_{t+1} 
= \frac{1-\beta_1}{1-\beta_1^{t+1}}(\beta^t_1 G_0 + \beta^{t-1}_1 G_1 + \dots G_t),
\]
and therefore
\[
\begin{split}
\tilde{U}_{t+1} 
& \overset{\cref{eq:91054}}\approx \beta_2^{t+1} U_0 + (1-\beta_2) \sum_{j=1}^{t+1} \beta_2^{t+1-j} \tilde{M}_j^2 \\
& = \beta_2^{t+1}U_0  \\ 
& + (1-\beta_2) \sum_{j=1}^{t+1} \beta_2^{t+1-j} \left( \frac{1-\beta_1}{1-\beta_1^j}  \right)^2 \left( \sum_{i=0}^{j-1} \beta_1^{j-1-i} G_i \right)^2.
\end{split}
\]
Expanding the rightmost side of the above, we find that the coefficient of $G_t^2$ is $(1-\beta_2) \left(\frac{1-\beta_1}{1-\beta_1^t}\right)^2$. 
This means $G_t^2$ contributes roughly $(1-\beta_2)\left( 1-\beta_1 \right)^2$ of the overall information in approximating the second moment, provided that $t$ is sufficiently large.
Recall that, in Adam, the term $G_t^2$ contributes $1-\beta_2^\text{Adam}$ of the overall second moment estimate for some $\beta_2=\beta_2^\text{Adam}$; see the second case in \cref{eq:adam-adaptation-rule}.
Hence, we need to keep $1-\beta_2^\text{Adam}$ in Adam identical to $(1-\beta_2)(1-\beta_1)^2$ in Alada in order to align their behavior. 
For a more concrete example, consider the default setting $\beta_1^\text{Adam}=0.9,\beta_2^\text{Adam}=0.999$ in Adam.
We can choose $\beta_1=0.9$ in Alada and solve for $1-\beta_2 = (1-0.999)(1-0.9)^2$, yielding $\beta_2=0.9$.
Therefore, we recommend setting $\beta_1=\beta_2=0.9$ in Alada; our experiment suggests that this works well in most cases.

\subsection{Extension to tensors}
We extend the alternating adaptation method to operate on tensor-shaped  variables.
Suppose a  variable $Y$ is an order-$\tau$ tensor of dimension $k_1\times \dots \times k_\tau$.
We propose reshaping $Y$ into a matrix whose width and height are as close as possible.
Precisely, consider solving the problem
\begin{equation}\label{eq:reshape-tensor}
j_* = \argmin \limits_{j} \left|\prod_{i=1}^j k_i - \prod_{i=j+1}^\tau k_i\right|,
\end{equation}
and set 
\[
m = \prod_{i=1}^{j_*} k_i \text{ and } n = \prod_{i=j_*+1}^\tau k_i.
\]
Then, we reshape $Y$ into a matrix of size $m\times n$.
This results in $m \approx n$ and hence $m+n \ll mn = \prod_{i=1}^\tau k_i$, ensuring that we do benefit from the memory-reduction effect of Alada.
On the other hand, the reshaping can be achieved by first flattening entries of $Y$ from dimension $(j_*+1)$ to dimension $\tau$ and then flattening from the first dimension to dimension $j_*$.
On platform where the tensor memory is in the same order as the dimensions (such as PyTorch), no real data copying is required. 
For example, we can simply use ``\texttt{Y.view(m, n)}'' on PyTorch without introducing memory overheads.
In Alada, we apply the reshaping operation to the term $V_{t+1}$, and the subsequent procedures are performed as if on a matrix optimization problem.
Similarly, when updating the decision variable (i.e., in line 21 of \Cref{alg:Alada-formal}), we reshape the terms $\tilde{U}_{t+1}$ and $\tilde{M}_{t+1}$ to have the same shape as $X_t$. Again, this will not cause data copying.

\section{Convergence}
We verify the performance of Alada in terms of first-order optimality.
Below are assumptions used in the analysis, all of which are customary for adaptive stochastic optimization algorithms.

\begin{assumption}\label{assumption:objective-value-lower-bound}
	The objective value is lower bounded by some constant $f_* \in \mathbb{R}$, i.e., 
	\[
		f(X) \ge f_*,\qquad \forall X.
	\]
\end{assumption}
\begin{assumption}\label{assumption:unbiased-and-bounded-sampling}
	The gradient sampling is unbiased such that
	\begin{equation}\label{eq:unbiasedness-gradient-sampling}
		\mathbb{E}_{\xi \sim \mathcal{D}}[\nabla F(X;\xi)] = \nabla f(X),\qquad \forall X.
	\end{equation}
	In addition, the $\ell_\infty$ norm of the stochastic gradient is upper bounded by some constant $G\in \mathbb{R}_+$, i.e., 
	\begin{equation}\label{eq:gradient-norm-bound}
		\|\nabla F(X;\xi)\|_\infty\le G,\qquad \forall X,\xi.
	\end{equation}
\end{assumption}
\begin{assumption}\label{assumption:L-smoothness}
	The objective function is $L$-smooth such that
	\begin{equation}\label{eq:L-smoothness}
		\|\nabla f(X') - \nabla f(X)\| \le L\|X'-X\|, \qquad \forall X,X'.
	\end{equation}
\end{assumption}

The following gives the convergence rate of Alada in approaching a first-order optimal solution.

\begin{theorem}\label{theorem:main}
	Consider solving problem \cref{eq:definition} with \Cref{alg:Alada-formal}. Choose the step-size as
	\begin{equation}\label{eq:step-size-setting}
			\eta_t = \eta(1-\beta_1^{t+1})
	\end{equation}
	for some positive $\eta$.
	Suppose \Cref{assumption:objective-value-lower-bound,assumption:unbiased-and-bounded-sampling,assumption:L-smoothness} hold.
	Then, the iterations generated satisfy 
	\begin{equation}\label{eq:convergence-bound}
		\frac{1}{T}\sum_{t=0}^{T-1}\mathbb{E}[\|\nabla f(X_t)\|^2] 
		\le \frac{2\Gamma\Delta_f}{\eta T} 
		+ \frac{LmnG^2\Gamma}{\epsilon} \eta
		+ \frac{\Gamma^2\Phi}{\epsilon}  
	\end{equation}
	where 
	\[
	\Delta_f = f(X_0) - f_*,
	\]
	\begin{equation}\label{eq:Gamma}
		\Gamma = \sqrt{\frac{2\sqrt{mn}G^2 + \epsilon}{1-\beta_2}},
	\end{equation}
	and
	\begin{equation} \label{eq:PHI}
		\Phi = 	mnG^2 \left(\frac{8}{T} \frac{1}{1-\beta_1} + \frac{\eta^2}{(1-\beta_1)^2} \frac{2L^2}{\epsilon} + 1-\beta_1 \right).
	\end{equation}
\end{theorem}

The above bound consists of three parts.
The first part is related to the initial setting (due to the $\Delta_f$ term); it can be reduced by using a large step-size.
The second part is introduced by the error when performing updates, and it diminishes when $\eta \to 0$.
The third part is caused by the first moment estimation, as it is where the decay parameter $\beta_1$ is involved.
All three parts involve the decay parameter $\beta_2$. 
It is then possible to investigate the effectiveness of the two momenta.
\begin{remark}[Impact of momenta]
	Examining the asymptotic bound in \cref{eq:convergence-bound}, we observe that the decay parameter $\beta_1$ has a non-linear impact on the performance. 
	Increasing $\beta_1$ may slow down the convergence (if the second term dominates the bound), but can improve the optimality of the best solution found during the search (as the third term can be made small). 
	On the other hand, increasing $\beta_2$ seems to degrade the performance consistently.
	This is probably because the obtained bound is not tight enough. 
	The numerical experiments presented later show that the recommended setting $\beta_1=0.9$ substantially improves the performance compared to the setting $\beta_1=0$ (i.e., when the first moment estimation is disabled).
	However, the impact of $\beta_2$ is shown to be negligible compared to that of $\beta_1$.
\end{remark}

Below we consider two commonly used step-size settings, which trade convergence rate for optimality.

\begin{corollary}[$T$-independent step-size]\label{corollary:T-independent-step-size}
Consider the same setting as in \Cref{theorem:main}.
Suppose $\beta_1$ and $\beta_2$ are constant.
Choose any $\epsilon \ge 2\sqrt{mn}G^2$ and specify the step-size as $\eta = \frac{(1-\beta_1)^{1.5}}{L} \sqrt{\frac{\epsilon}{2}}$. Then we have
\begin{equation*}
\begin{split}
	& \frac{1}{T}\sum_{t=0}^{T-1}\mathbb{E}[\|\nabla f(X_t)\|^2] 
	\le 5mnG^2\frac{1-\beta_1}{1-\beta_2} \\ 
  & \qquad\qquad + \frac{1}{T}\left( \frac{4L\Delta_f}{(1-\beta_1)^{1.5}\sqrt{1-\beta_2}} + \frac{16mnG^2}{(1-\beta_1)(1-\beta_2)} \right)
\end{split}
\end{equation*}
\end{corollary}

\begin{remark}[Comparison to Adam]
	When $\beta_1,\beta_2$ are constant, Alada converges at a rate of $\mathcal{O}(1/T)$ to a suboptimal solution. 
	This rate aligns with that achieved by Adam~\cite{defossez_simple_2022}, suggesting that the reduction in memory overheads of Alada does not harm its theoretical performance. 
\end{remark}
We note that within the above setting, solutions obtained might be suboptimal.
One can close the suboptimal gap with a $T$-dependent hyperparameter setting, at the price of a slower convergence rate.

\begin{corollary}[$T$-dependent $\eta$]\label{corollary:T-dependent-step-size}
Let $\beta_1 = 1 - T^{-\frac{1}{2.5}}$.
In the same setting as in \Cref{theorem:main,corollary:T-independent-step-size}, we have 
\begin{equation*}
	\frac{1}{T}\sum_{t=0}^{T-1}\mathbb{E}[\|\nabla f(X_t)\|^2]
	= \left( \frac{4L\Delta_f}{\sqrt{1-\beta_2}} + \frac{21mnG^2}{1-\beta_2} \right) T^{-\frac{1}{2.5}}.
\end{equation*}
\end{corollary}

\begin{remark}[Iteration complexity]
	For sufficiently small $\delta>0$, Alada finds a solution with gradient norm smaller than $\delta$ in at most $\mathcal{O}(\frac{1}{\delta^5})$ iterations.
	This is worse than that reported in~\cite{defossez_simple_2022}. We speculate this is the side effect of using first moment estimates in estimating the second moment. However, our experiments suggest that Alada performs similarly to Adam when using the default $\beta_1$ setting.
\end{remark}


\section{Numerical study}
We verify the performance of Alada on three natural language processing tasks, namely language understanding, neural machine translation, and language modeling.
\subsection{Generic settings}
We choose Adam~\cite{kingma_adam:_2015} and its memory-efficient variant Adafactor~\cite{shazeer_adafactor_2018} for comparison.
The decay parameters in Adam are set to $\beta_1=0.9$ and $\beta_2=0.999$. Correspondingly, we use $\beta_1=\beta_2=0.9$ in Alada, following our suggestions in \Cref{ss:decay-parameter-settings}.
In Adafactor, the first moment estimation is disabled, and the decay parameter for the second moment estimation is set to $\beta_2 = 0.999$. The regularization coefficient $\epsilon$ in both Adam and Adafactor is fixed to $10^{-8}$. We choose $\epsilon = 10^{-16}$ in Alada\footnote{The preconditioner in Adam is of the form $\left(\sqrt{U}+\epsilon\right)^{-1}$ while that in Alada is $\left(\sqrt{U+\epsilon}\right)^{-1}$, where $U$ is some second moment estimate. Therefore, the regularization coefficients in all considered algorithms are roughly in the same magnitude.}.
All algorithms are equipped with a diminishing step-size scheme $\eta_t = \eta_0/(1 - t/T)$, where $T$ is the total number of iterations and the initial step $\eta_0$ is tuned for each test problem.
All algorithms are run on a single A800 GPU.

\subsection{Language understanding}
We fine-tune two pretrained language models, namely BERT-Base~\cite{devlin_bert_2019} and OPT-1.3B~\cite{zhang_opt_2022}, for natural language understanding tasks. 
Seven datasets from the GLUE benchmark suite~\cite{wang_glue_2019} are considered.
All algorithms are assigned with a computation budget of 3 epochs with a batch size (``bsz'') of 32.
The initial learning rate $\eta_0$ is tuned in $10^{-5} \times \{1, 2, 4\}$ for BERT-Base and $10^{-5} \times \{1/4, 1/2, 1\}$ for OPT-1.3B. Every algorithm is run three times independently on each task and with each $\eta_0$. 

\begin{figure}[tb] 
	\centering
	\subfloat{\includegraphics[width=0.4\textwidth]{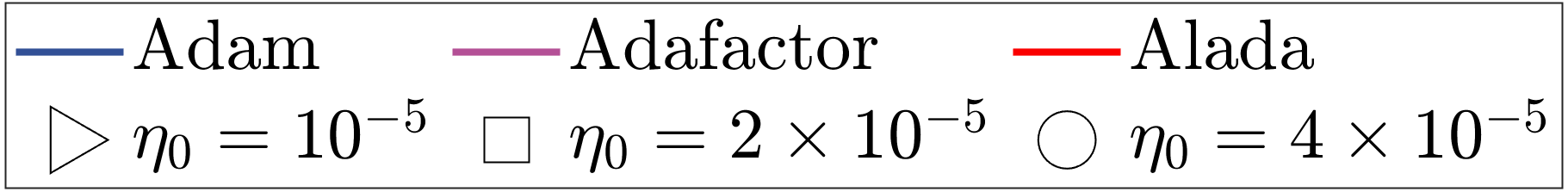}} \\[-2ex]
	\addtocounter{subfigure}{-1}
	\subfloat[COLA]{\includegraphics[width=0.24\textwidth]{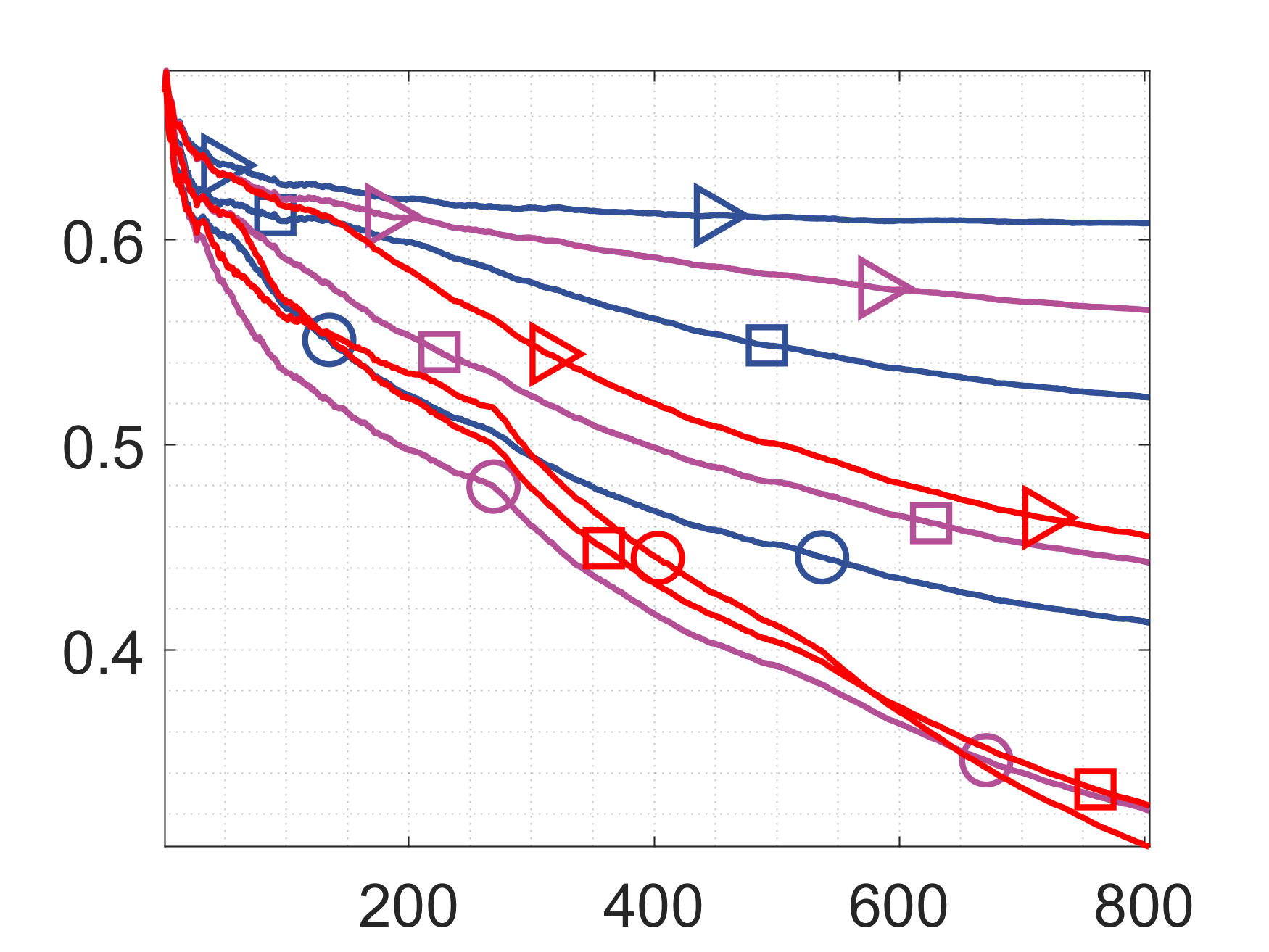}} 
	\subfloat[MNLI]{\includegraphics[width=0.24\textwidth]{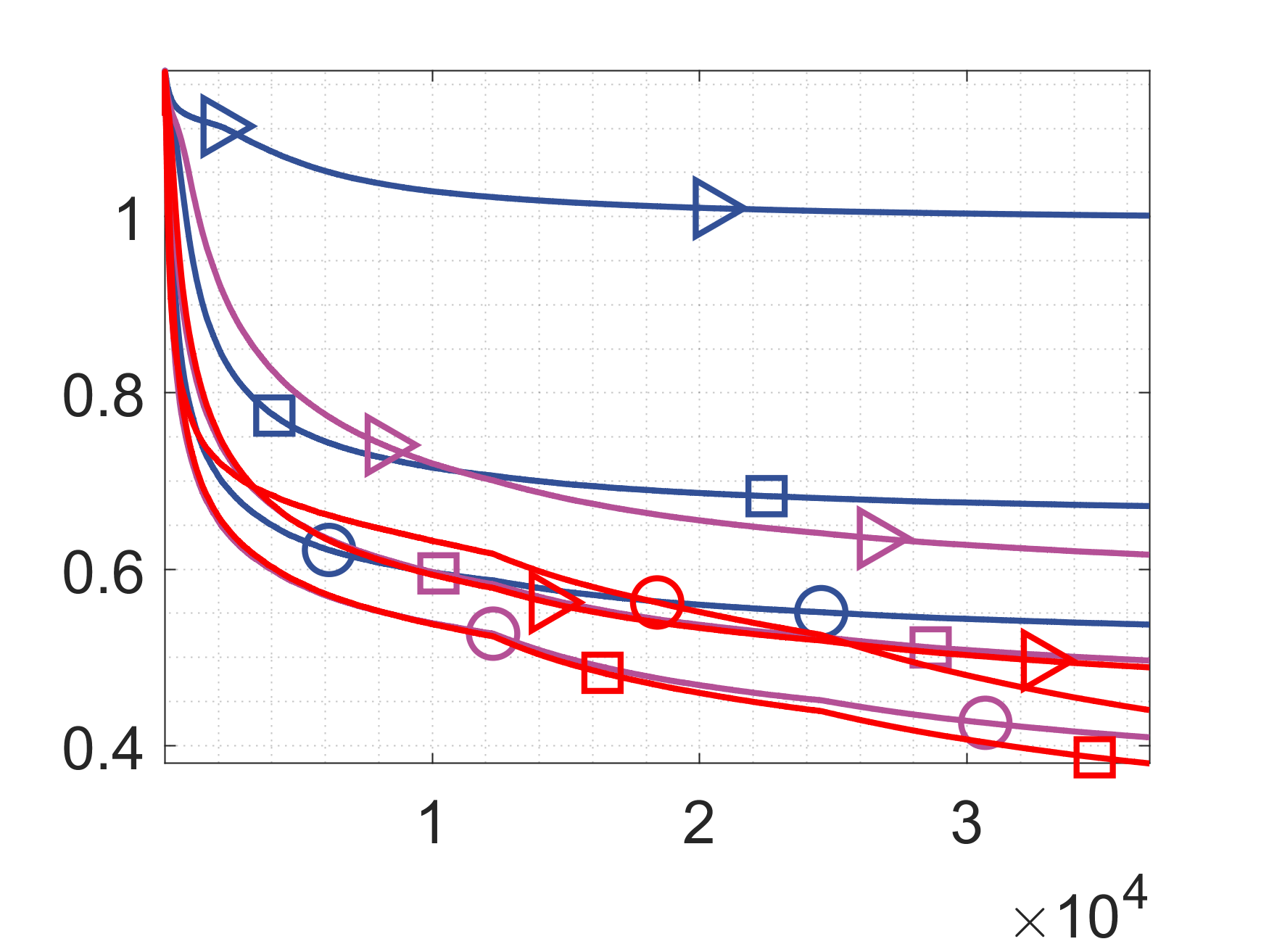}} \\
	\subfloat[MRPC]{\includegraphics[width=0.24\textwidth]{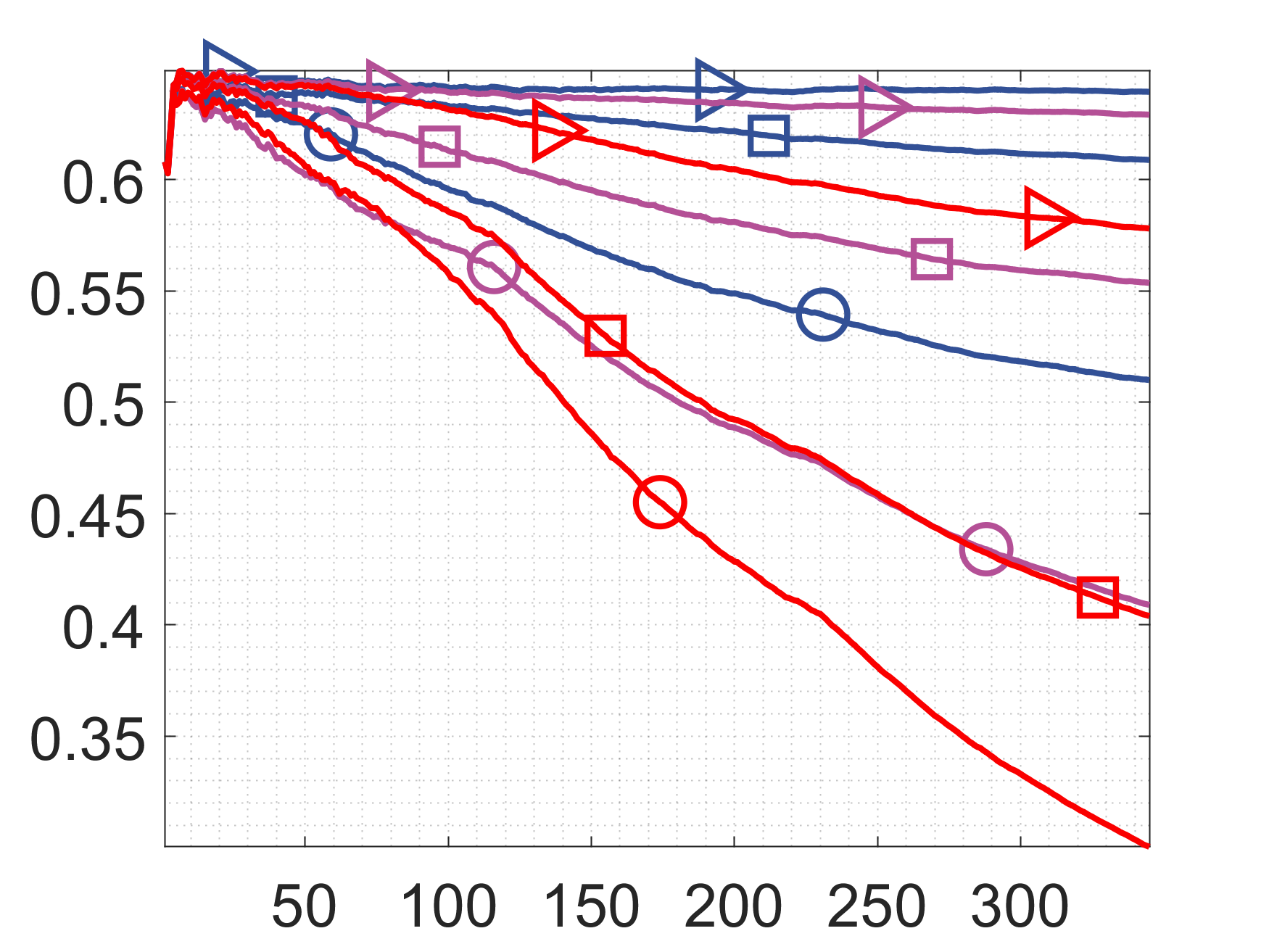}}
	\subfloat[QQP]{\includegraphics[width=0.24\textwidth]{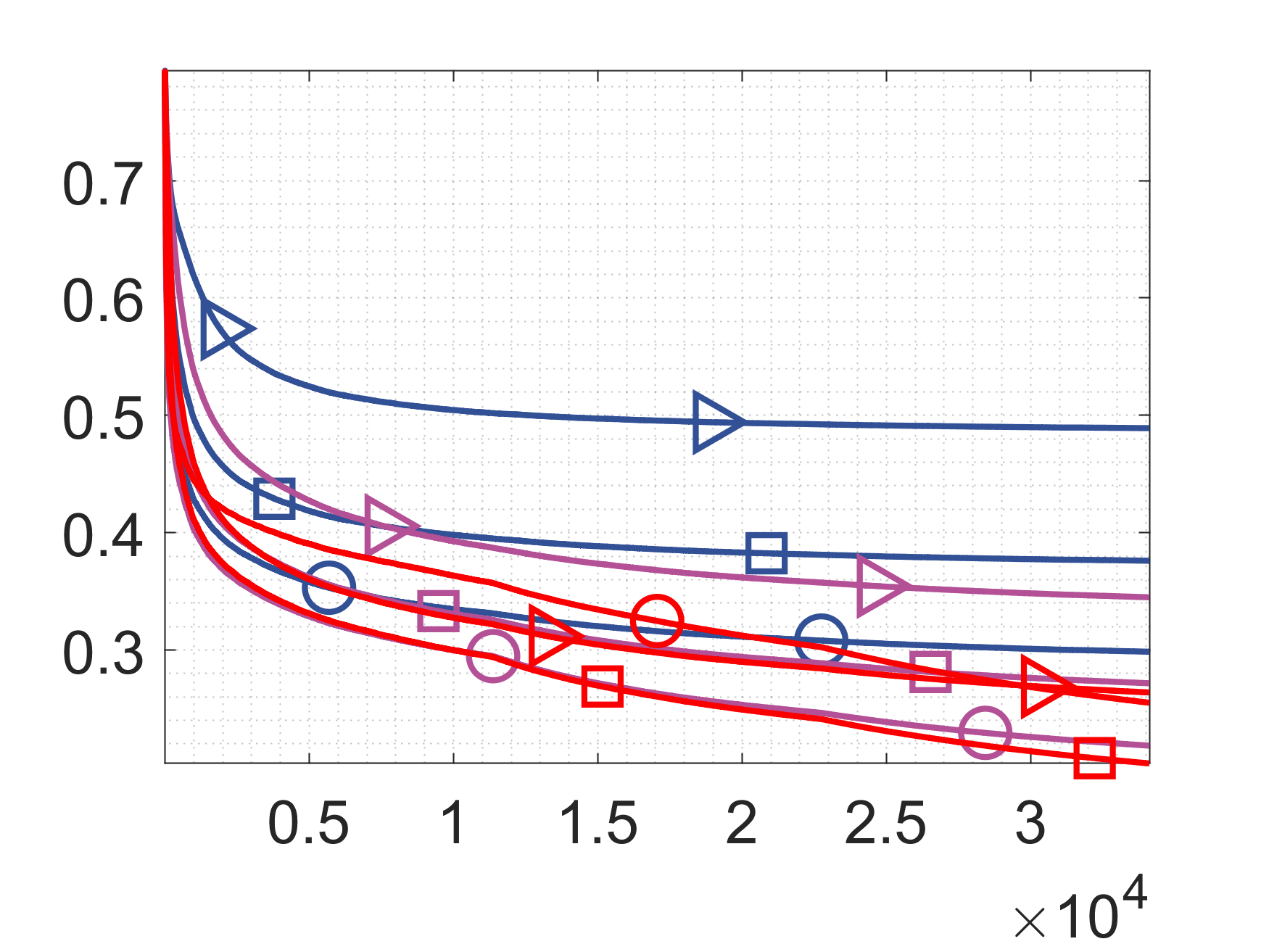}} \\
	\subfloat[QNLI]{\includegraphics[width=0.24\textwidth]{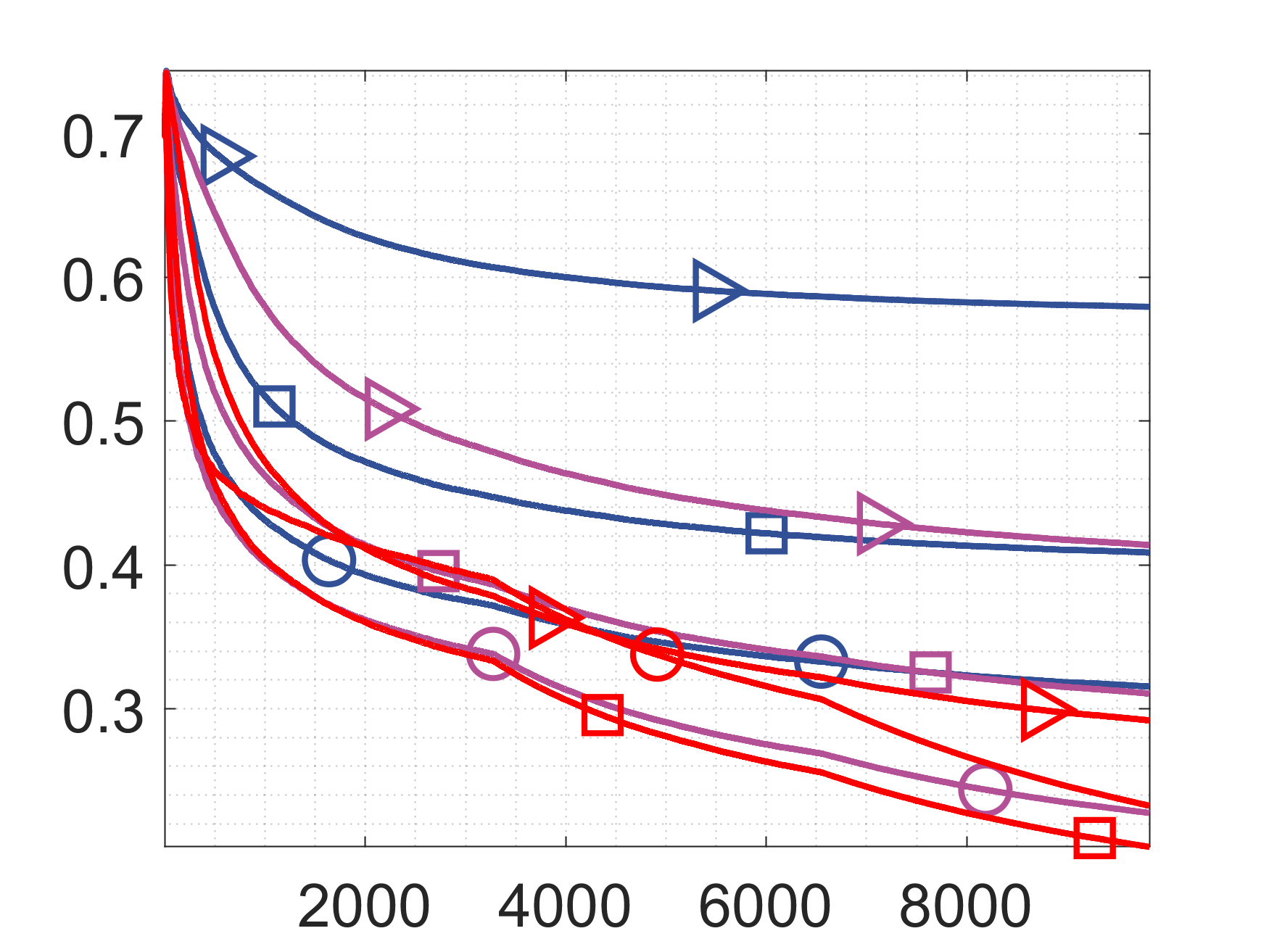}} 
	\subfloat[RTE]{\includegraphics[width=0.24\textwidth]{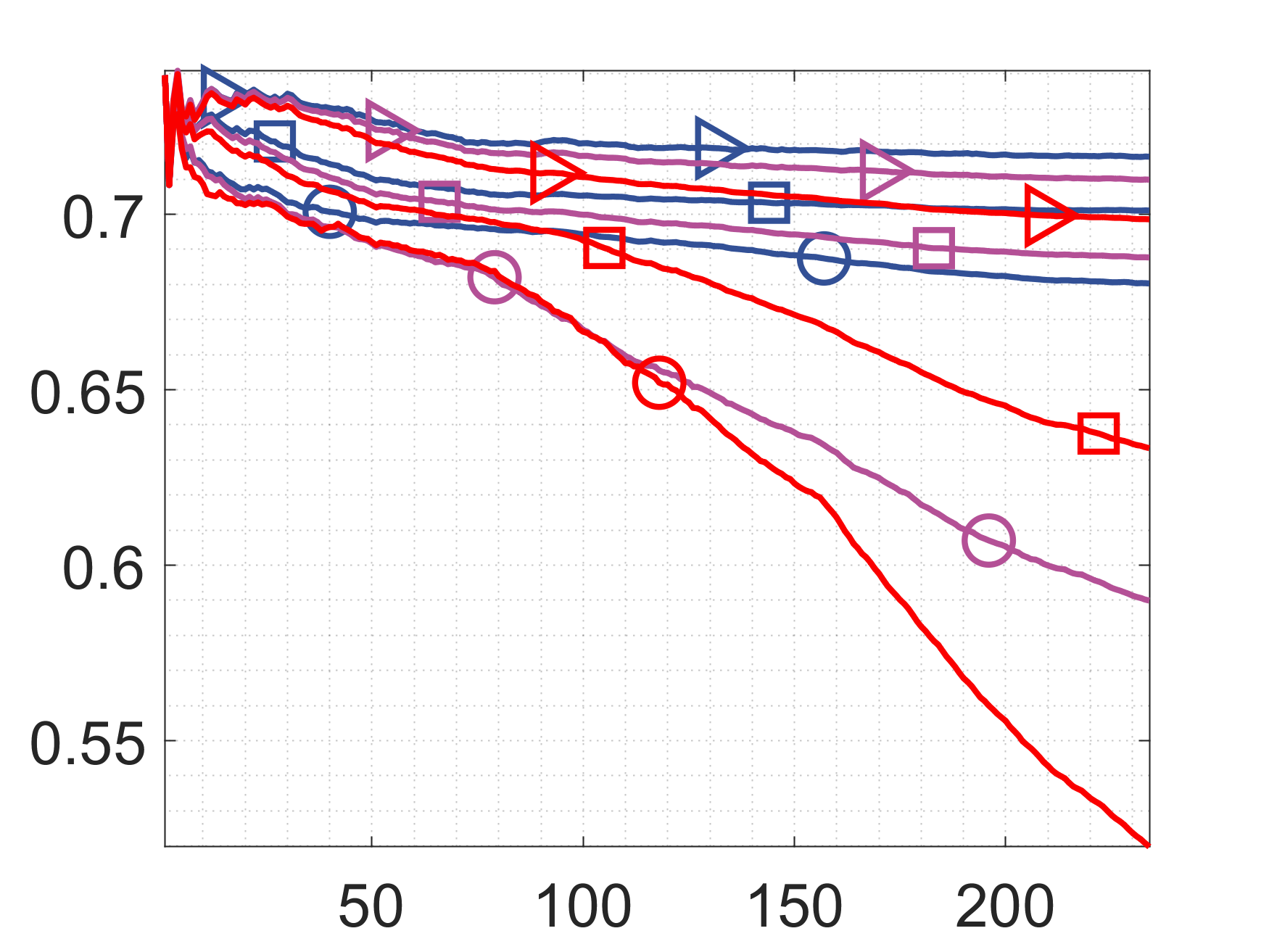}} \\
	\subfloat[SST2]{\includegraphics[width=0.24\textwidth]{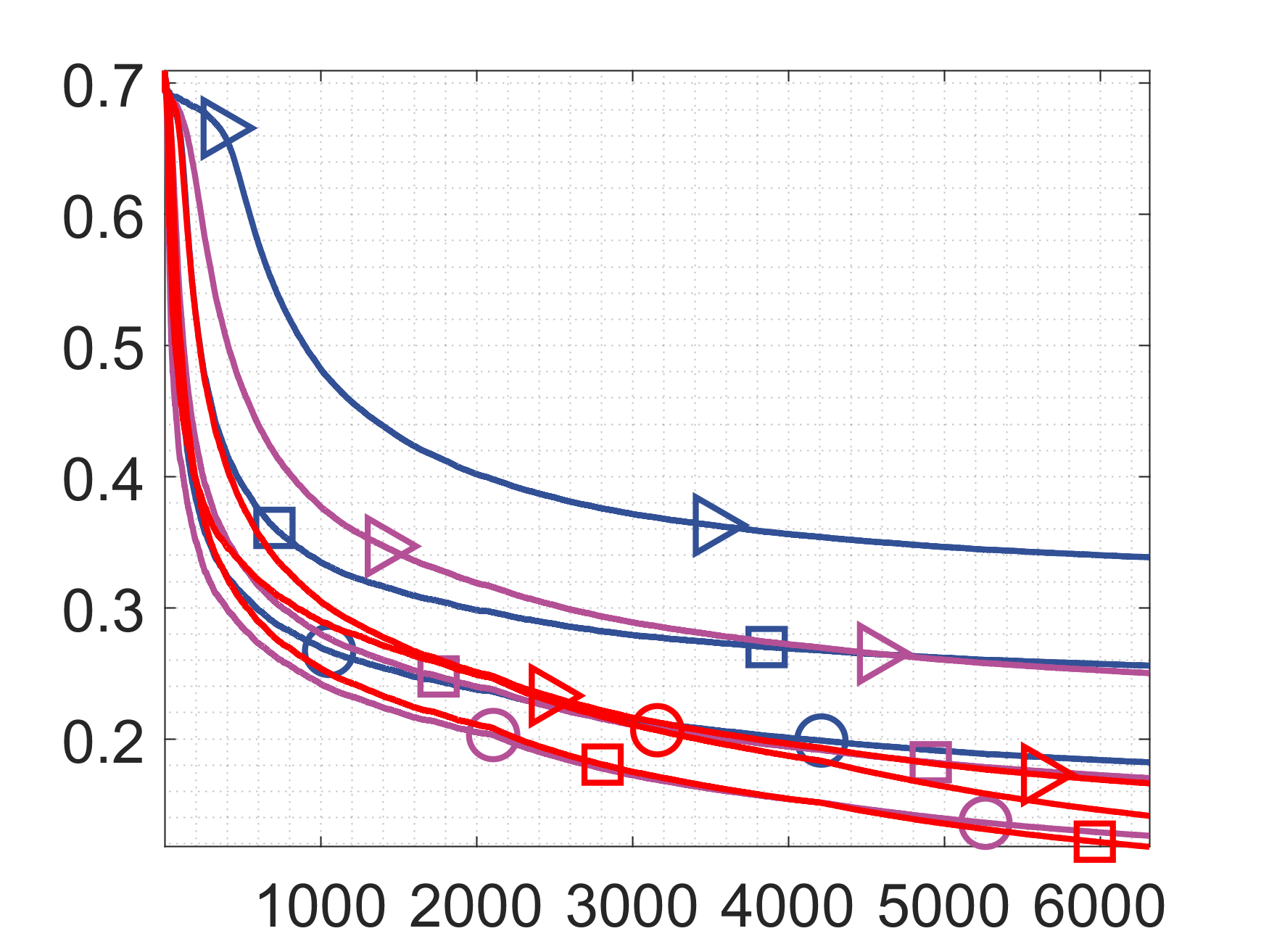}}
	\caption{Convergence trajectories of Adam, Adafactor, and Alada when fine-tuning the BERT-Base model on GLUE tasks. The $x$-axis denotes the number of update steps. The $y$-axis denotes the cumulative average of training losses.}
	\label{fig:training-trajectories-BERT-GLUE}
\end{figure}

We first show in \Cref{fig:training-trajectories-BERT-GLUE} the training performance evaluated on the BERT-Base model. Both Alada and Adafactor are competitive with Adam on the majority of the tasks. Alada achieves faster convergence than Adafactor on MRPC and RTE while performing similarly to Adafactor on the rest tasks. 
The results are similar when evaluated on OPT-1.3B and are given in the supplement.

\Cref{tab:GLUE-metric} gives the mean test metric values evaluated on the test samples on each benchmark task. 
We report F1 values for MRPC and QQP, Matthews correlation coefficients for COLA and accuracies for the rest.
We observe that Alada is competitive with Adam and Adafactor in terms of the generalization ability. 
Compared to those obtained with Adam, metrics obtained with Alada are seen to improve on all tasks.
Alada also obtains slightly higher average metrics than Adafactor on both models.
\begin{table}[htbp]
	\centering
	\caption{Mean metrics of GLUE benchmark problems evaluated on BERT-Base and OPT-1.3B}
	\label{tab:GLUE-metric}
  \scriptsize
\begin{tabular}{lcccccccc}
	\toprule
	& COLA  & MNLI  & MRPC  & QQP   & QNLI  & RTE   & SST2 \\
	\midrule
	& \multicolumn{7}{c}{BERT-Base} \\
	\midrule
Adam  & 58.71 & 83.16 & 89.44 & 85.86 & 91.24 & 66.78 & 92.39 \\
Adafactor & \textbf{60.46} & 83.48 & \textbf{90.98} & 87.28 & 91.20 & 66.90 & 92.24 \\
Alada & 58.11 & \textbf{84.53} & 90.19 & \textbf{87.71} & \textbf{91.71} & \textbf{68.11} & \textbf{93.08} \\
	\midrule
	& \multicolumn{7}{c}{OPT-1.3B} \\
	\midrule
Adam  & 58.53 & 85.37 & 82.52 & 85.38 & 90.59 & 64.86 & 95.14 \\
Adafactor & 64.06 & 87.02 & 86.75 & 87.42 & 92.34 & 74.12 & \textbf{95.98} \\
Alada & \textbf{64.56} & \textbf{87.99} & \textbf{89.09} & \textbf{88.69} & \textbf{92.38} & \textbf{78.82} & 95.68 \\
\bottomrule
\end{tabular}%
\end{table}

\subsection{Neural machine translation}
We fine-tune the T5-Small model~\cite{raffel_exploring_2020} to translate German, Czech, Russian, Romanian, Finnish, and Turkish to English, respectively. 
The model is pretrained and involves 60M parameters. 
All datasets are from the WMT16 benchmark set.  
All algorithms run for 10 epochs with a batch size of 64.
The step-size $\eta_0$ is chosen from $10^{-3} \times \{1, 2, 4, 8\}$. 
We report the mean results over five independent runs.

\begin{figure}[htb] 
	\centering
	\subfloat{\includegraphics[width=0.47\textwidth]{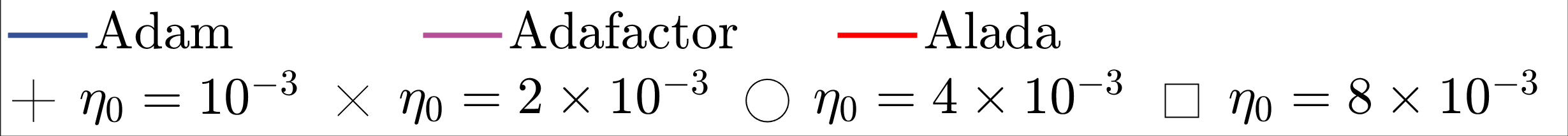}} \\[-2ex]
	\addtocounter{subfigure}{-1}
	\subfloat[De-En]{\includegraphics[width=0.24\textwidth]{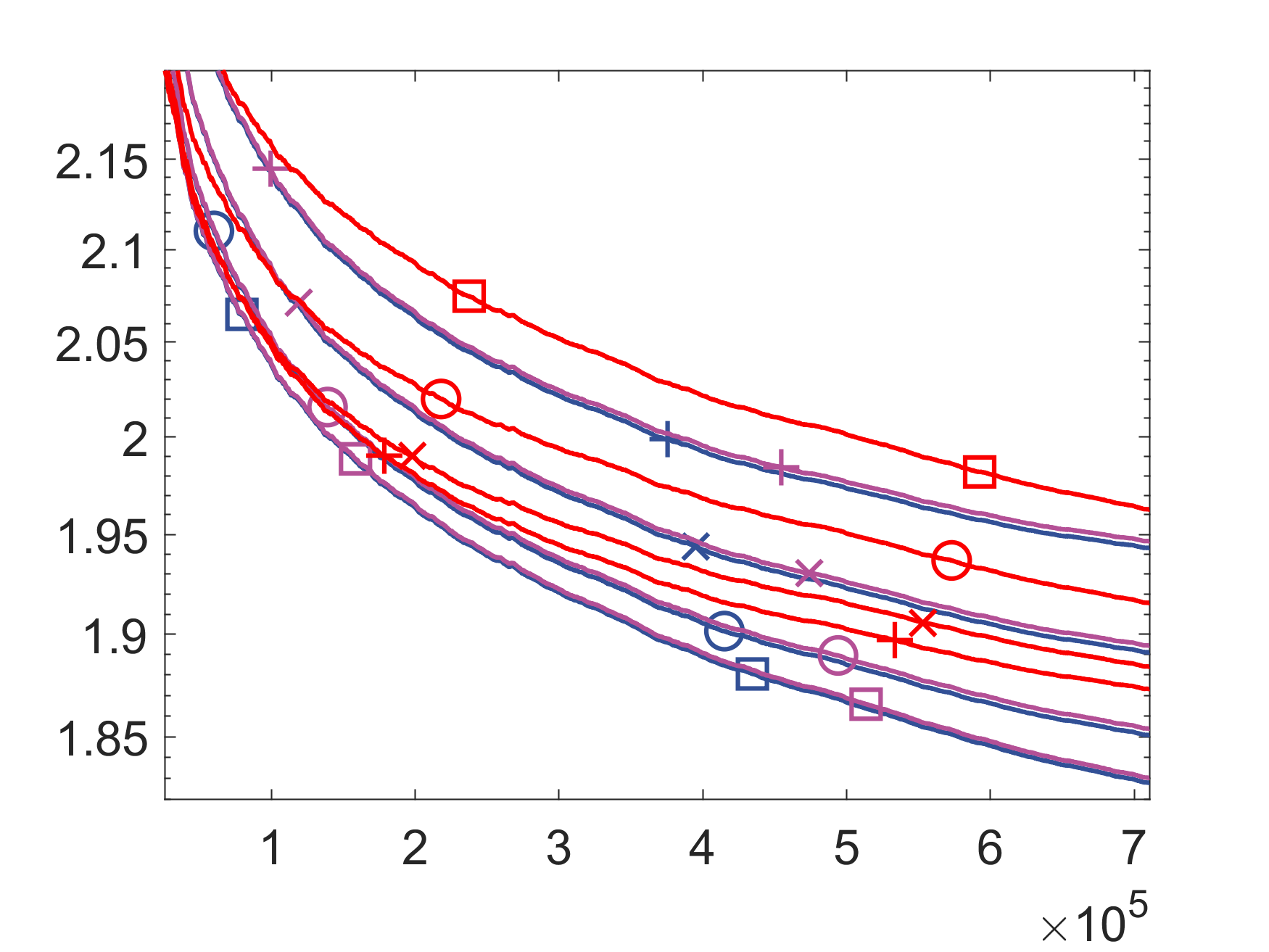}}
	\subfloat[Cs-En]{\includegraphics[width=0.24\textwidth]{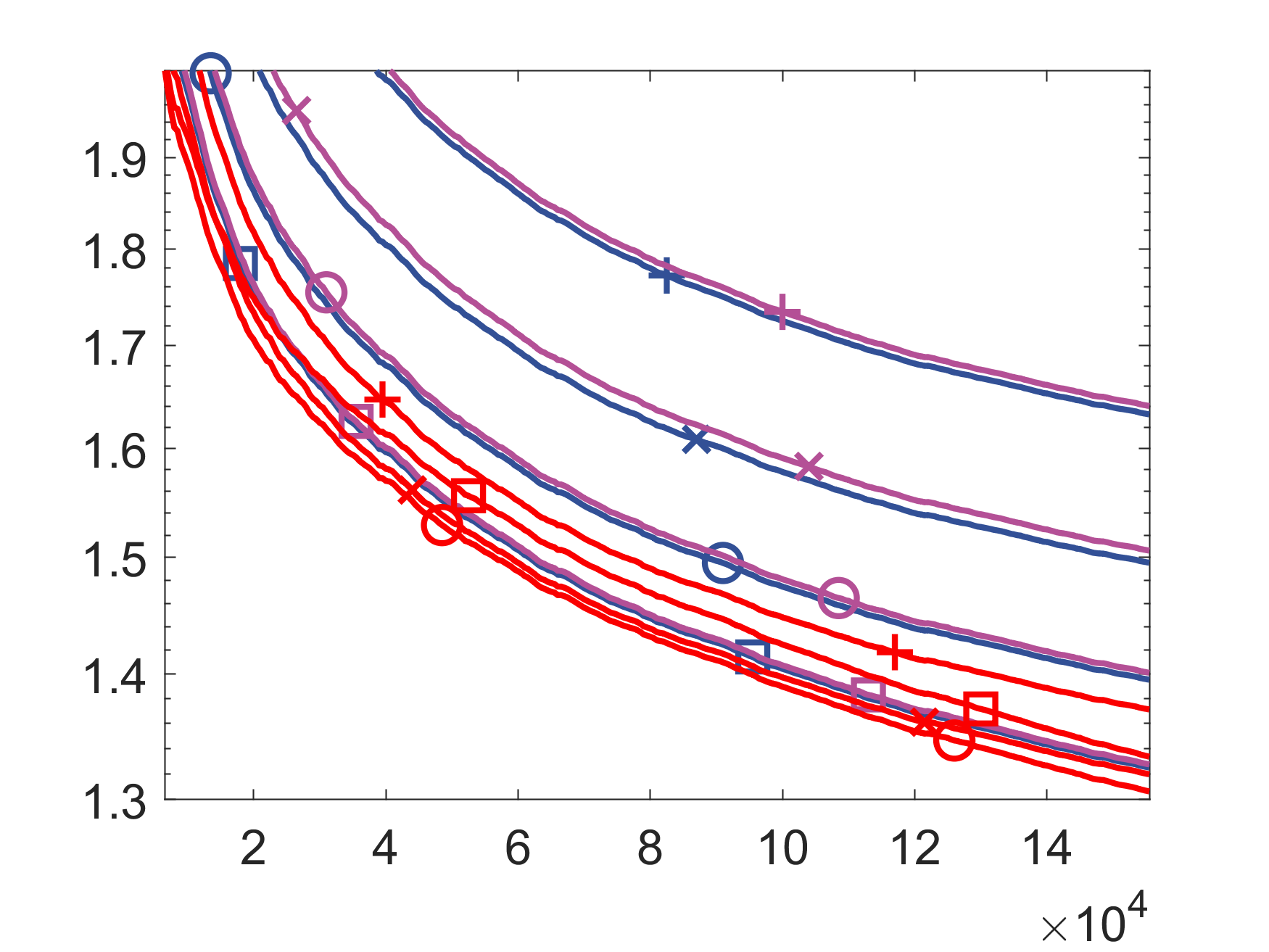}} \\
	\subfloat[Ru-En]{\includegraphics[width=0.24\textwidth]{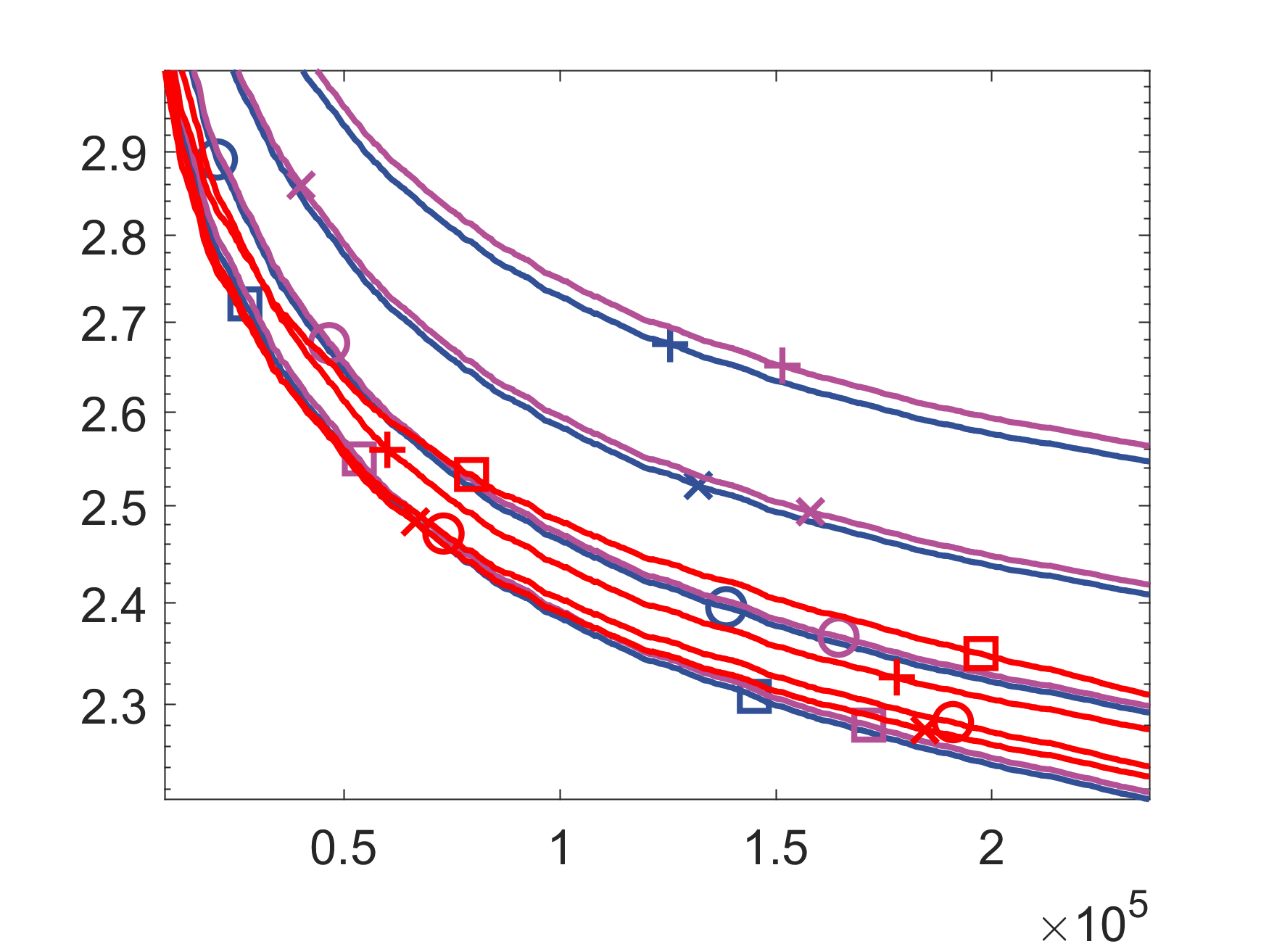}} 
	\subfloat[Ro-En]{\includegraphics[width=0.24\textwidth]{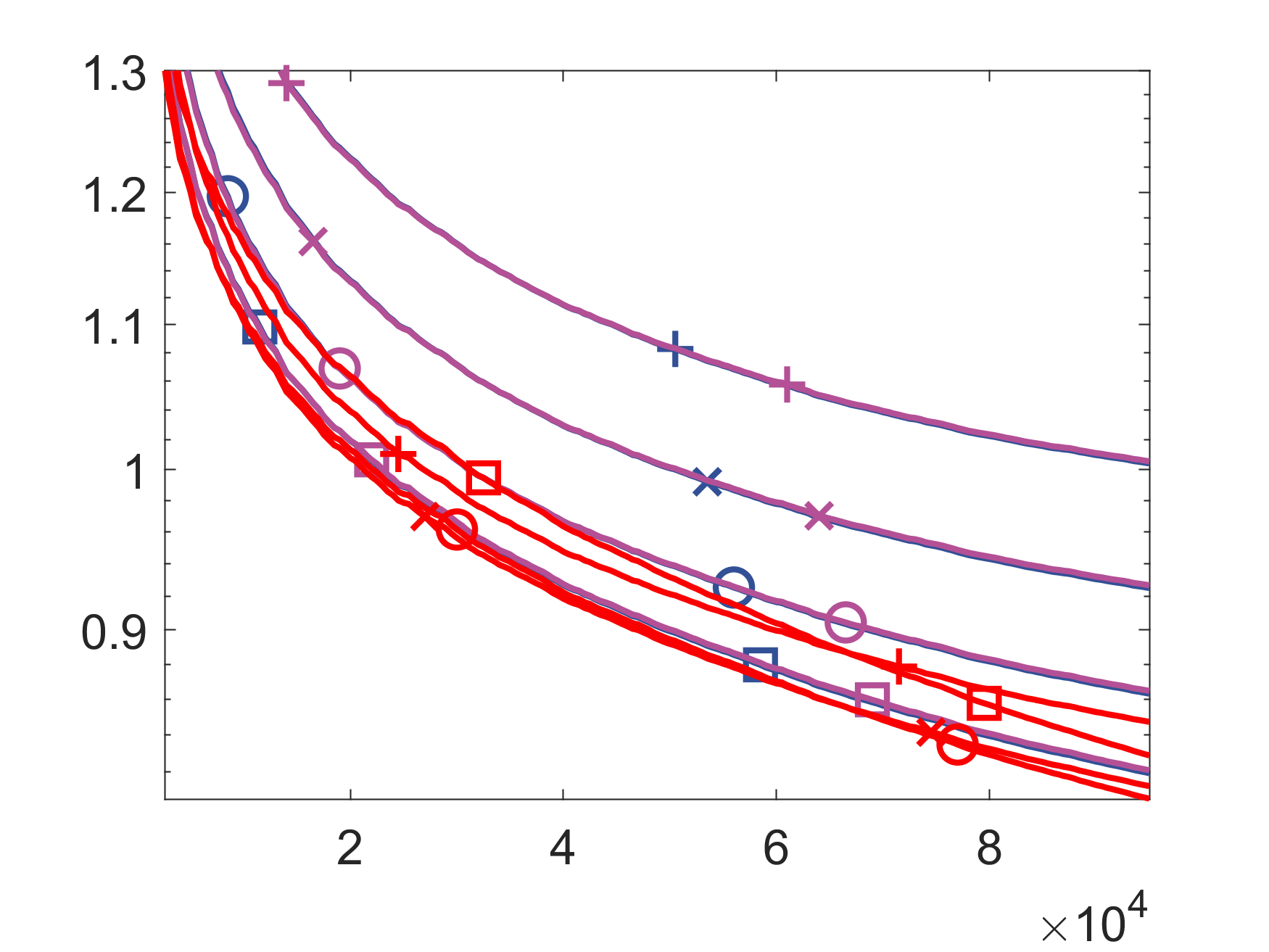}} \\
	\subfloat[Fi-En]{\includegraphics[width=0.24\textwidth]{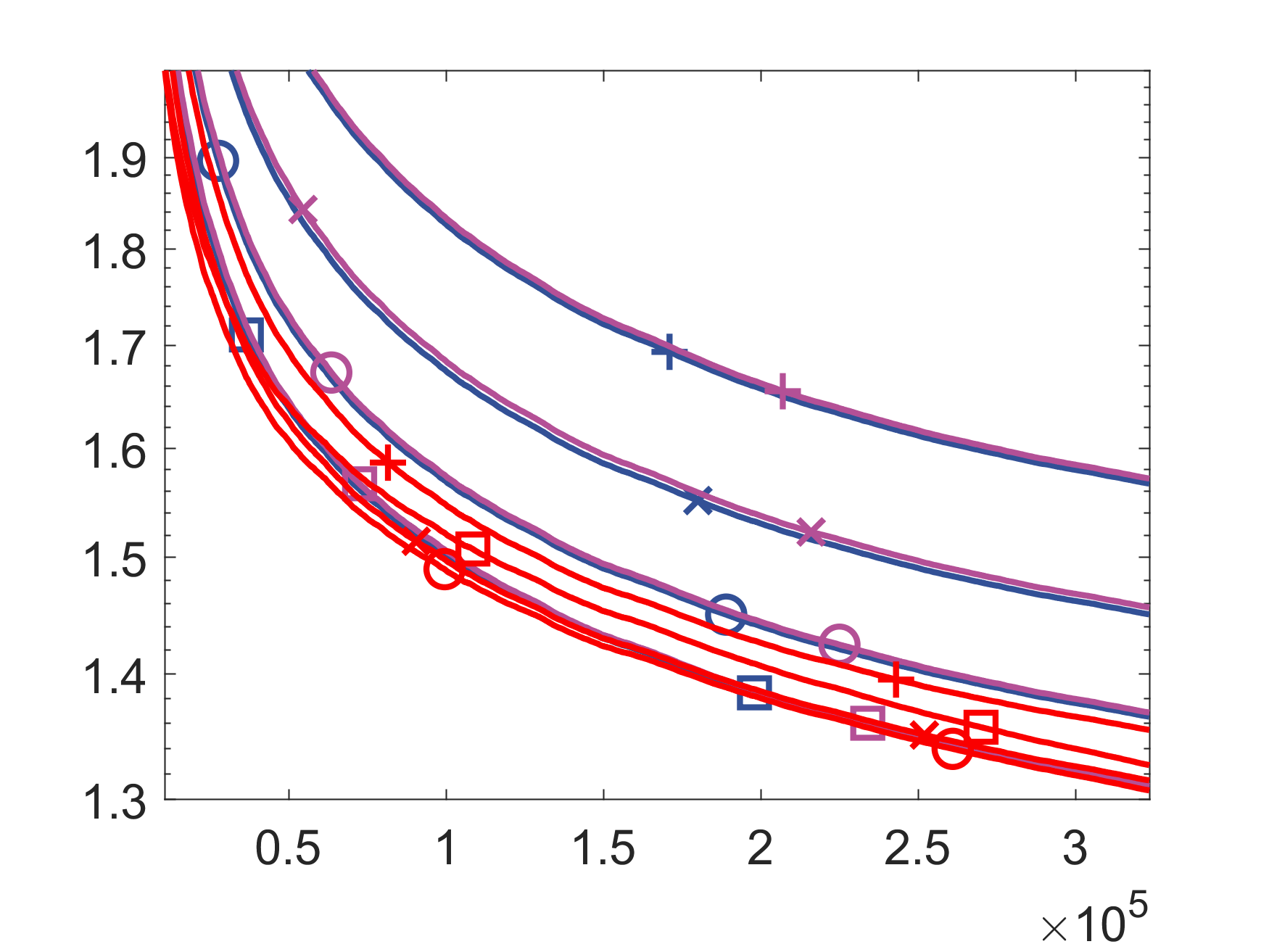}}
	\subfloat[Tr-En]{\includegraphics[width=0.24\textwidth]{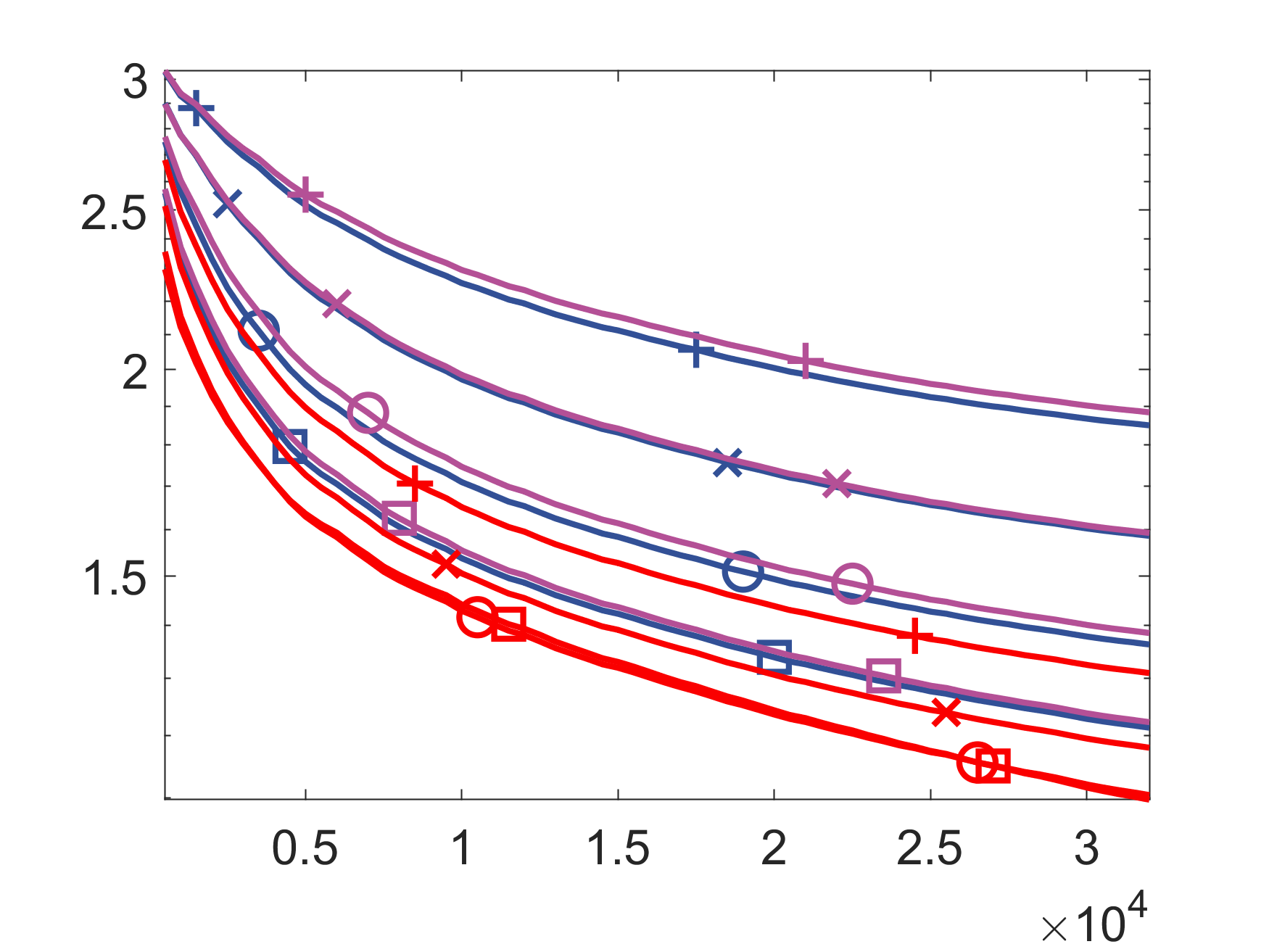}}
	\caption{Convergence trajectories of Adam, Adafactor, and Alada when fine-tuning T5-Small for neural machine translation. The $x$-axis denotes the number of update steps. The $y$-axis denotes the cumulative average of training losses.}
	\label{fig:training-trajectories-WMT16}
\end{figure}

\Cref{fig:training-trajectories-WMT16} plots the trajectories of stochastic training losses.
Alada achieves similar training performance compared to Adam and Adafactor, while being more robust to different step-sizes compared.
\Cref{tab:neural_translation} summarizes the highest BLEU~\cite{post_call_2018} scores achieved by each algorithm.
It is observed that Alada performs the best on five out of the six tasks, demonstrating its competitive generalization ability.

\begin{table}[htbp]
	\centering
	\caption{Mean BLEU metrics of neural machine translation}
	\label{tab:neural_translation}
\begin{tabular}{lcccccc}
	\toprule
	& De-En & Cs-En & Ru-En & Ro-En & Fi-En & Tr-En\\
	\midrule
Adam &  30.831 & 23.452 & 23.634 & \textbf{36.246} & 19.831 & 18.920\\
Adafactor &  30.745 & 23.419 & 23.554 & 36.219 & 19.889 & 19.015\\
Alada &  \textbf{30.936} & \textbf{24.113} & \textbf{24.042} & 36.160 & \textbf{20.204} & \textbf{20.242} \\
\bottomrule
\end{tabular}%
\end{table}

\subsection{Language modeling}
We perform language modeling tasks on the WikiText2~\cite{merity_pointer_2017} dataset.
Two pretrained models from the GPT2 family are considered, namely GPT2-Small with 124M parameters and GPT2-XL with 1.5B parameters.
We run Adam, Adafactor, and Alada with a budget of 20 epochs. 
We group the tokens from the raw dataset into sequences of fixed length 1024. 
For GPT2-Small, we use a batch size of 24 in all algorithms, and we tune the initial learning rate in $10^{-5} \times \{1/2, 1, 2, 4, 8\}$.
For GPT2-XL, we set the batch size to 4 in Alada and Adafactor.
Adam fails to run with a batch size of 4 due its huge memory usage, so for Adam we set the batch size to 2.
The initial learning rate is tuned in $10^{-6} \times \{1/8, 1/4, 1/2, 1, 2\}$ for GPT2-XL.
All other algorithmic settings are kept the same as in the GLUE tasks.

\begin{figure}[tb] 
	\centering
  \scriptsize
	\subfloat[GPT2-Small, bsz=24]{\includegraphics[width=0.24\textwidth]{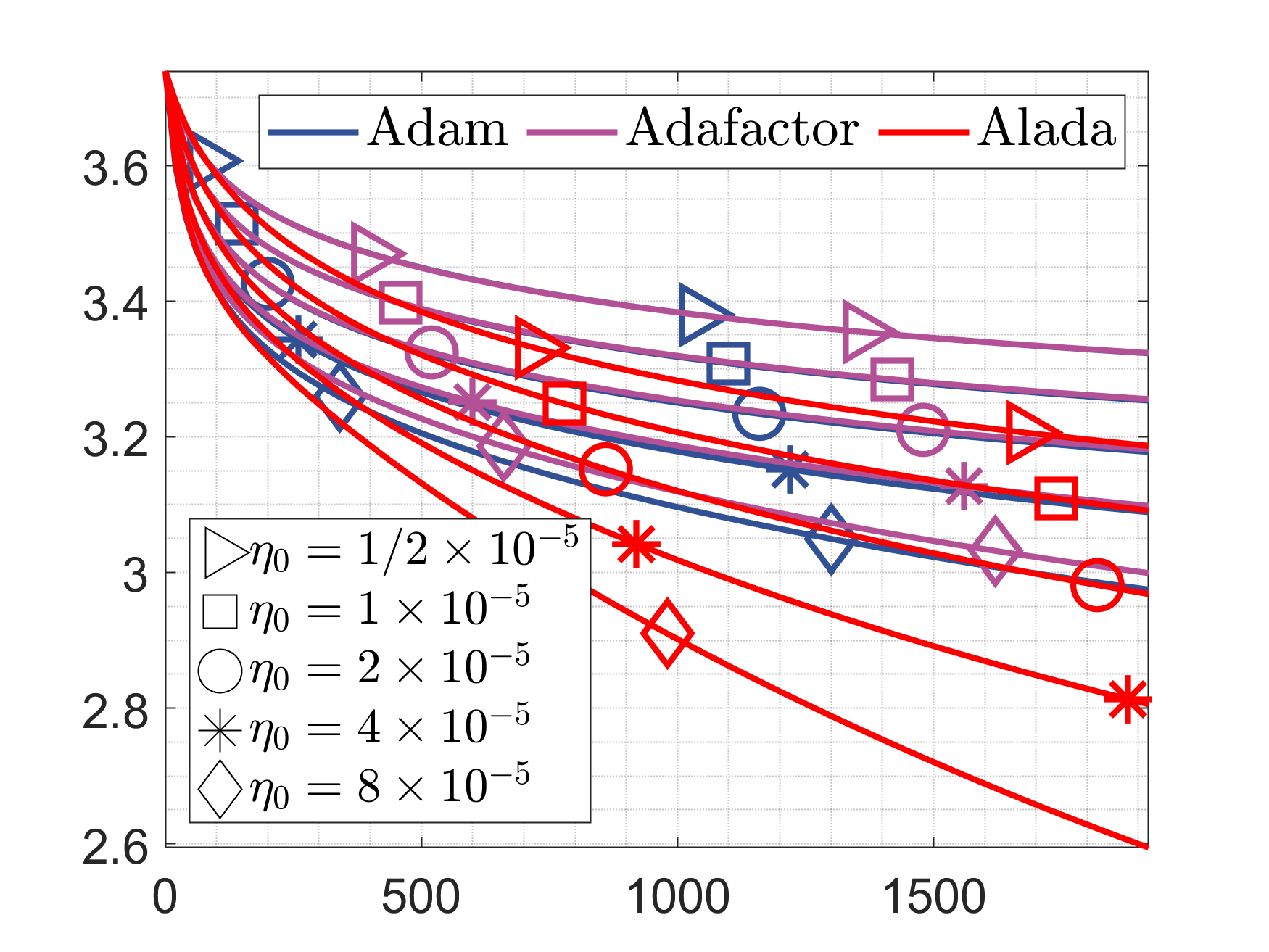}} 
	\subfloat[GPT2-XL, bsz=2]{\includegraphics[width=0.24\textwidth]{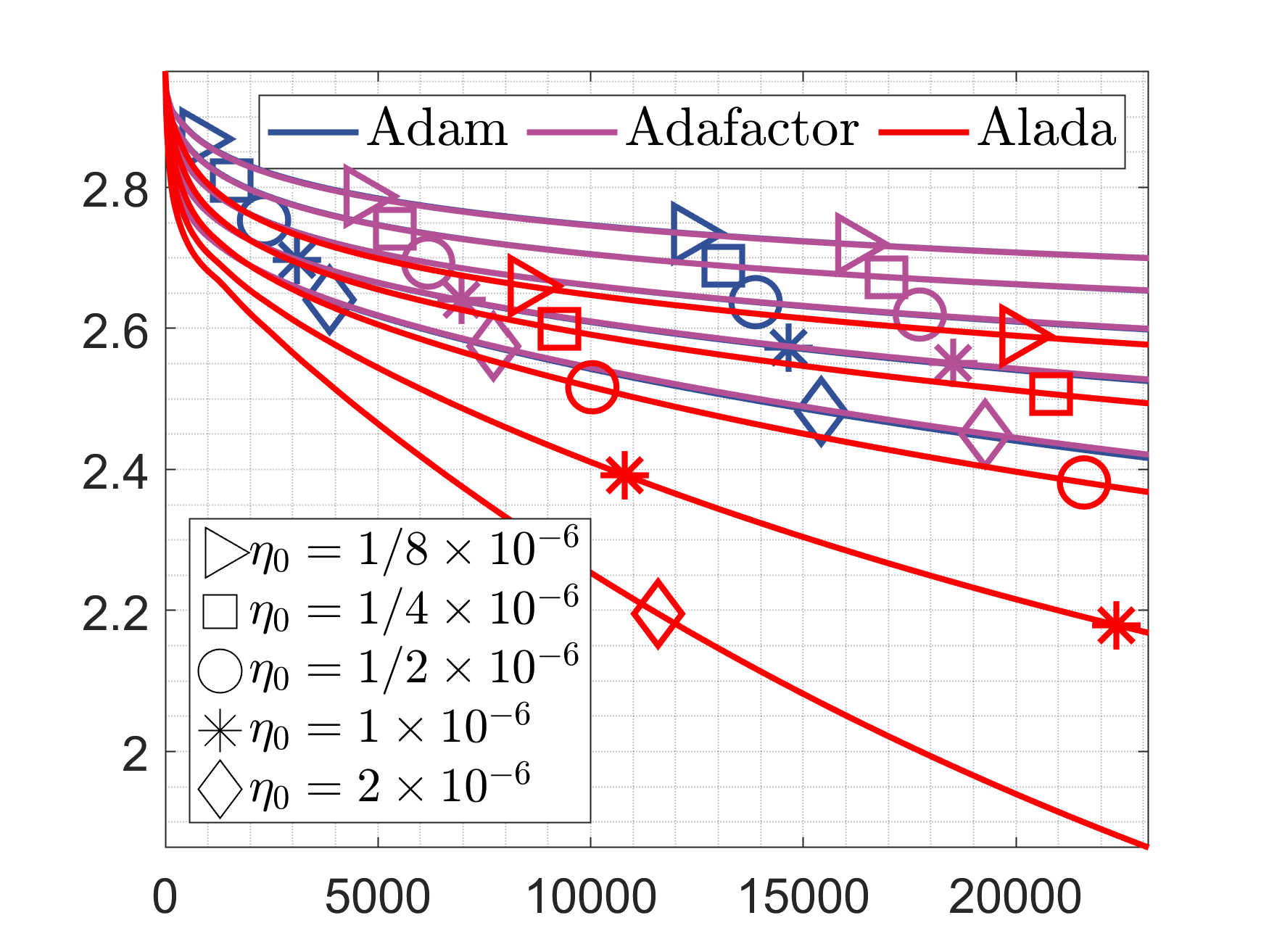}}\\
	\subfloat[GPT2-XL, bsz=4]{\includegraphics[width=0.24\textwidth]{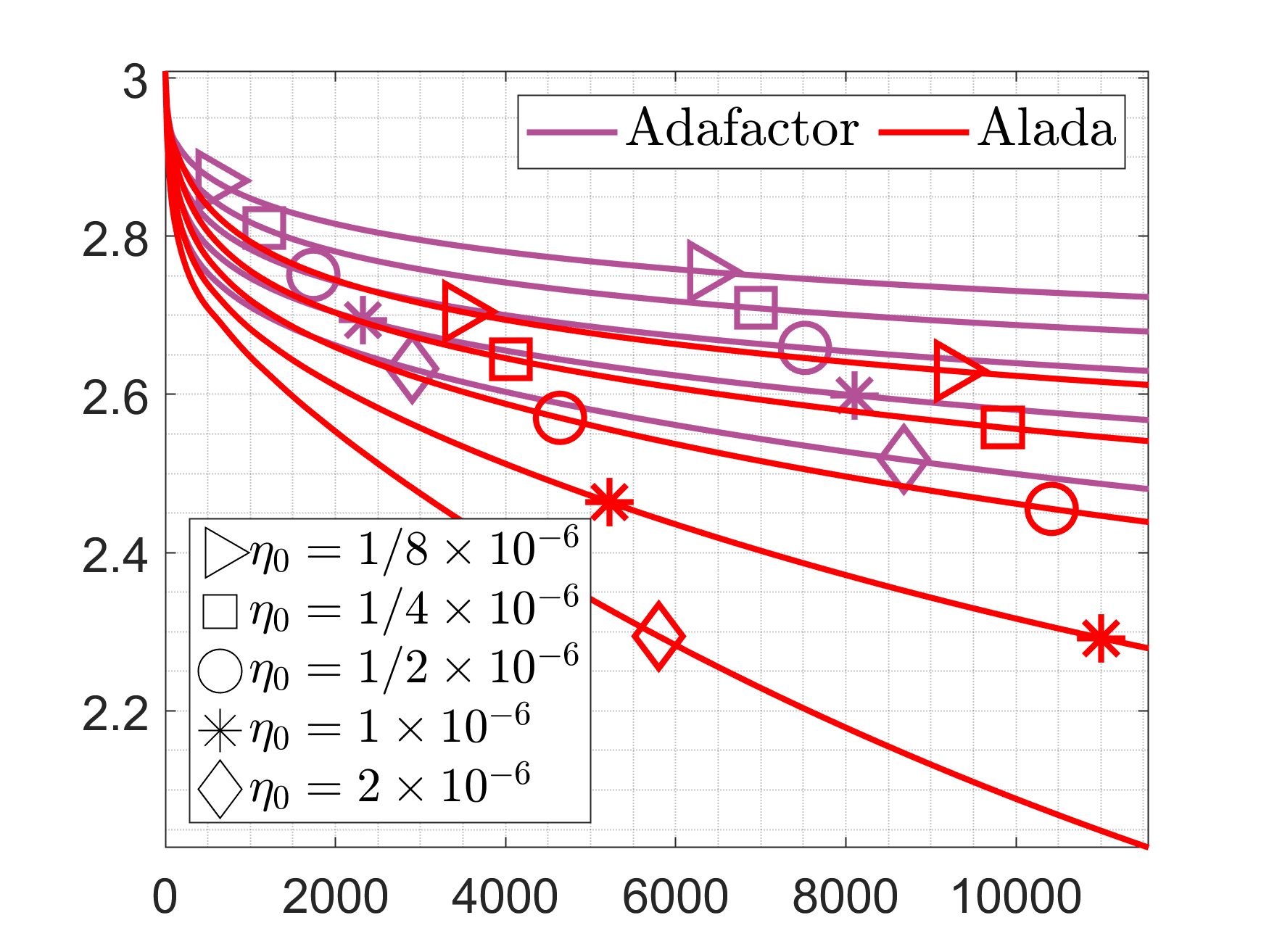}}
	\caption{Convergence trajectories of Adam, Adafactor, and Alada on the WikiText-2 language modeling task. The $x$-axis denotes the number of update steps. The $y$-axis denotes the cumulative average of training losses.}
	\label{fig:training-trajectories-GPT2-WikiText2}
\end{figure}

We plot the training curves of algorithms in \Cref{fig:training-trajectories-GPT2-WikiText2}.
Adafactor and Adam perform similarly in general while both are outperformed by Alada.
We also evaluate the perplexities on the test set. 
The results are given in \Cref{tab:LM-PPL-results}. 
Alada achieves the best perplexity among all algorithms.

\begin{table}[htbp]
	\centering
	\caption{Test perplexity evaluated on the WikiText-2 language modeling task.}
	\label{tab:LM-PPL-results}
	\begin{tabular}{ccccc}
		\toprule
		GPT2 size &  Batch size & Adam  & Adafactor  & Alada  \\
		\midrule
		Small &  24 & 20.880 & 20.881 & \textbf{20.865} \\
		\multirow{2}{*}{XL} &  2 & 12.633 & 12.633 & \textbf{12.610} \\
		 &  4 & N/A & 12.650 & \textbf{12.634} \\
		\bottomrule
\end{tabular}%
\end{table}

\subsection{Computation cost}
We now investigate the computation cost of the proposed method.
The experiment is set up on the language modeling and neural machine translation tasks. In the former case we consider fine-tuning both GPT2-Small and GPT2-XL on WikiText2 and in the latter case we fine-tune T5-Small to translate German to English.
We compute the wall-clock time averaged over the first 10,000 steps and the corresponding peak memory consumed.
To reduce the cost due to forward/backward passes, the batch size is fixed to 1.
In this way, the results mainly reflect the overheads caused by the algorithm rather than the model's activation memory.
Other settings are kept the same as those in the previous sections.
\Cref{tab:computational_cost} reports the results, where we observe Alada reduces more than 30\% of the memory demand of Adam, showing the memory efficiency of our proposal.
Alada requires 20\% more per-step clock time than Adam, which is due to the overheads in decomposing and reconstructing the second moment estimate. 
We note that Alada, compared to Adafactor, involves an additional estimate to the first moment of gradients while introducing no significant overheads.

\newcommand{\tabincell}[2]{\begin{tabular}{@{}#1@{}}#2\end{tabular}}  
\begin{table}[htbp]
	\centering
  \scriptsize
	\caption{Peak memory usage and per-iteration wall-clock time}
	\label{tab:computational_cost}
	\begin{threeparttable}
	\begin{tabular}{llccc}
		\toprule
		& Task & Adam  & Adafactor  & Alada  \\
		\midrule
		\multirow{3}{*}{\tabincell{c}{Peak memory usage \\ (GB)}} & GPT2-Small + LM  & 3.483 & 2.534 & 2.571 \\
		& GPT2-XL	+ LM & 34.278 & 22.388 & 22.386 \\
		& T5-Small + NMT & 1.605 & 1.155 & 1.130 \\
		\midrule
		\multirow{3}{*}{\tabincell{c}{Per-step clock time \\ (s)}} & GPT2-Small + LM  & 0.045 & 0.057 & 0.055 \\
		& GPT2-XL	+ LM & 0.424 & 0.479 & 0.523 \\
		& T5-Small + NMT & 0.026 & 0.035 & 0.032 \\
		\bottomrule
	\end{tabular}%
	\begin{tablenotes}
		\item{$\bullet$} LM: language modeling on WikiText2, NMT: neural machine translation for translating German to English.
	\end{tablenotes}
	\end{threeparttable}
\end{table}

\subsection{Hyperparameter sensitiveness}
\begin{figure}[tb] 
	\centering
	\subfloat[Cs-En]{\includegraphics[width=0.24\textwidth]{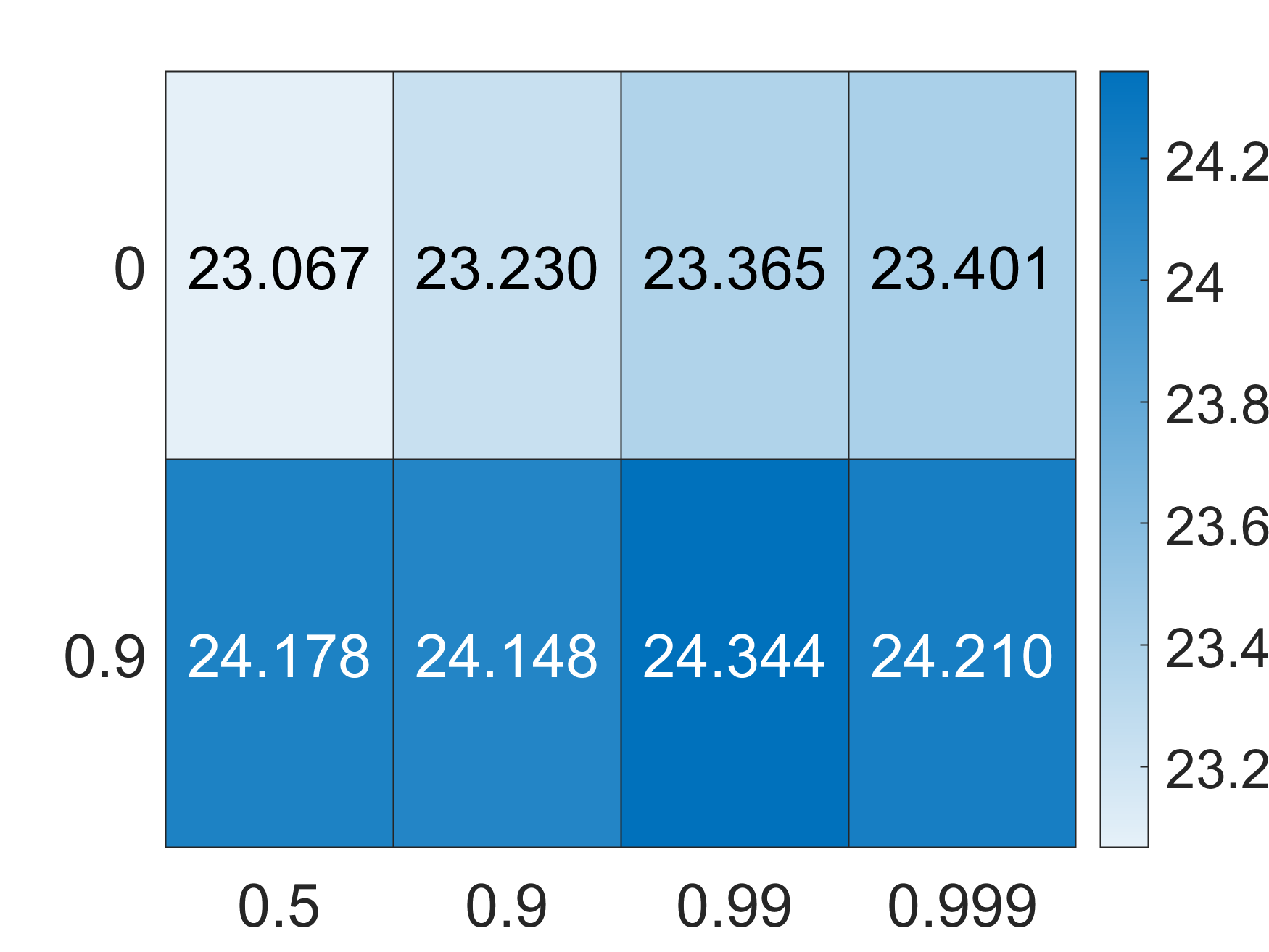}}
	\subfloat[Ro-En]{\includegraphics[width=0.24\textwidth]{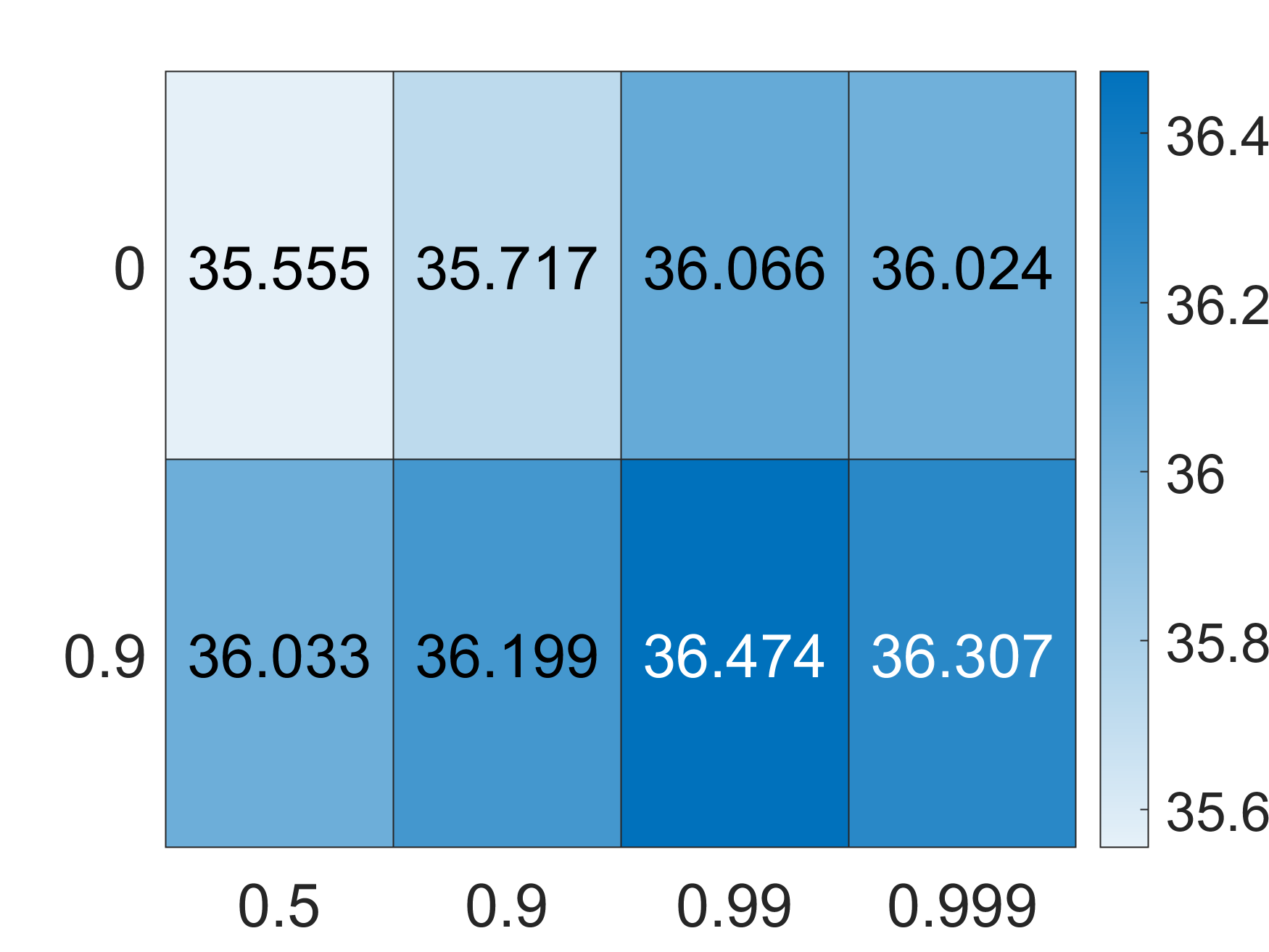}} \\
	\subfloat[Tr-En]{\includegraphics[width=0.24\textwidth]{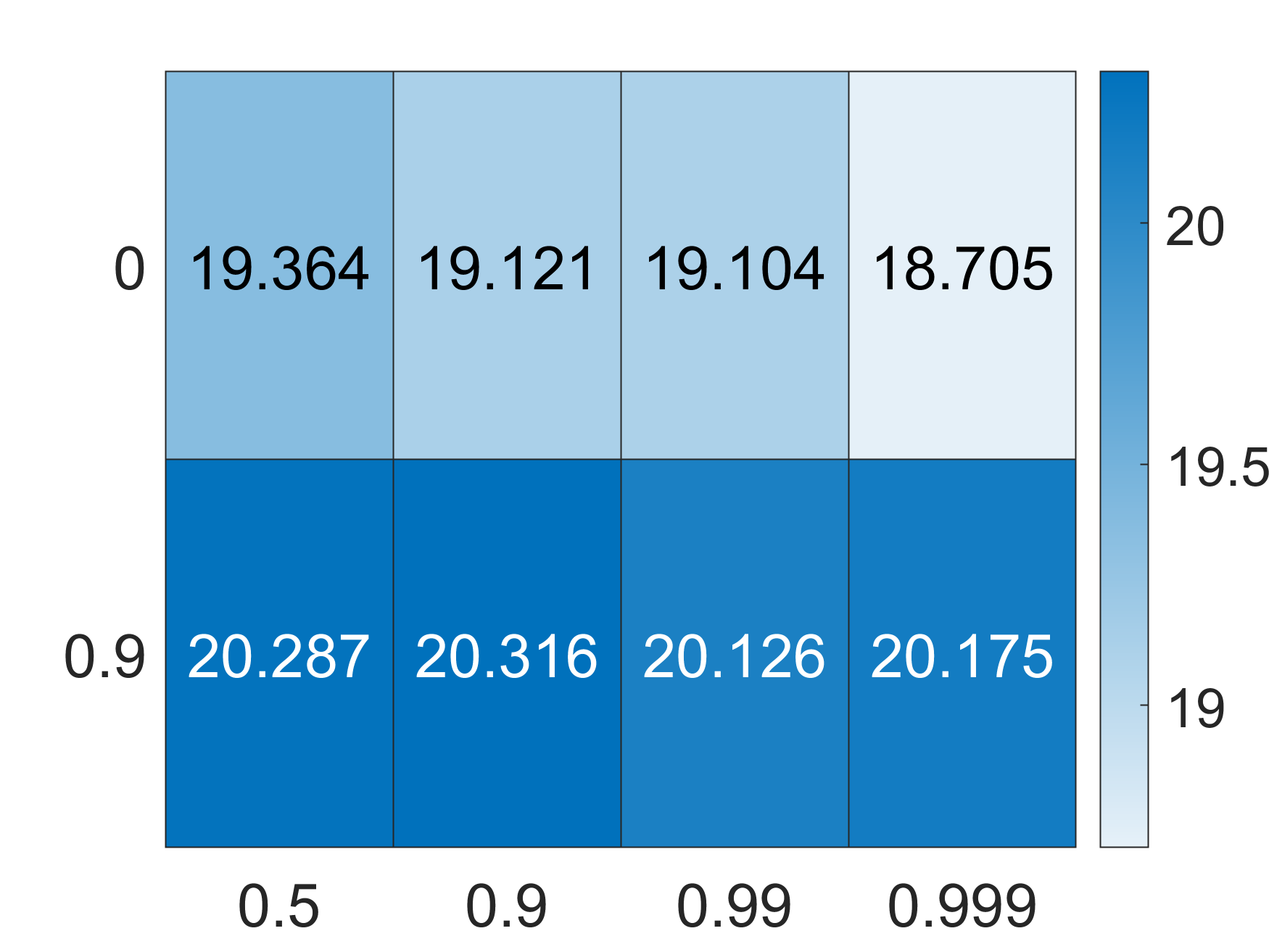}}
	\caption{Average BLEU metrics obtained by Alada on three machine translation tasks. 
	The $x$-axis and $y$-axis denote the $\beta_2$ and $\beta_1$ settings, respectively.
	The initial step-size $\eta_0$ is tuned.}
	\label{fig:parameter-sensitiveness-b1b2}
\end{figure}

Here we test the impact of $\beta_1, \beta_2$ on the performance of Alada.
We consider three neural machine translation tasks, namely Cs-En, Ro-En, and Tr-En.
We choose $\beta_1 \in \{0, 0.9\}$ and $\beta_2 \in \{0.5, 0.9, 0.99, 0.999\}$, and for each combination of $\beta_1$ and $\beta_2$ we tune the initial step-size $\eta_0$ in $10^{-4} \times \{1, 2, 4, 8\}$.
The algorithm is run three runs independently with each setting, and we report the mean BLEU values in \Cref{fig:parameter-sensitiveness-b1b2}. 
It is seen that $\beta_1 = 0.9$ is much better than $\beta_1=0$, demonstrating the necessity of using a momentum rule in estimating gradients.
The performance is, on the other hand, insensitive to the setting of $\beta_2$. 
Setting $\beta_2=0.9$ or 0.99 yields slightly better results, but the overall impact is insignificant.
Therefore, we recommend choosing $\beta_1=0.9$ and $\beta_2\in\{0.9,0.99\}$ by default.

\section{Related work}
Extensive efforts have been devoted to reducing the memory overheads of adaptive algorithms when solving problem \cref{eq:definition}. 
One popular approach, known as the low-rank adaptation (LoRA)~\cite{hu_lora_2022}, converts the original space $\mathbb{R}^{m\times n}$ into an affine subspace $X_0 + \mathbb{R}^{m\times r} \times \mathbb{R}^{r\times n}$, where $r\in \mathbb{Z}$ and $X_0\in\mathbb{R}^{m\times n}$ are pre-defined. 
This allows reducing the memory requirements of adaptive methods to $\mathcal{O}(rm+rn)$ memory.
If, in addition, the matrix $X_0$ is well initialized, i.e., there exist nearly optimal solutions in the reduced space, the performance degradation might be even negligible. 
This is usually the case of fine-tuning pretrained models, where $X_0$ is the pretrained model weights and of high-quality due to the model's good generalization ability~\cite{dettmers_qlora_2023,liu_multi-agent_2024,zhao_galore_2024}.
Another paradigm demonstrated success is to quantize the optimizer states to lower bitwidth~\cite{li_memory_2023,dettmers_8-bit_2021,dettmers_case_2023,dettmers_qlora_2023}.
This trades optimization stability and training accuracy for smaller memory footprints. 
The method proposed in this work is orthogonal to these approaches and can be used in conjunction with them.

Our proposed method belongs to the class of memory-efficient adaptive methods. 
Representative examples include Adafactor~\cite{shazeer_adafactor_2018}, CAME~\cite{luo_came_2023}, and SM3~\cite{anil_memory_2019}, all of which are variants of Adam and only differ in the way of modeling the second moment. 
Alada improves these methods in that Alada estimates the first moment of gradients without increasing memory usage.
An alternative way is to apply Adam to coordinate descent frameworks~\cite{beck_convergence_2013}.
Methods following this idea include \cite{lv_full_2024, pan_lisa_2024,zhao_galore_2024}. 
They typically divide the search procedure into several phases and at each phase a subset of decision variables are updated with Adam.
These methods are as memory-efficient as LoRA, but differ in that the iterations are not necessarily constrained in an affine subspace with fixed bases. 

Our idea of reusing first moment estimates in estimating the second moment was previously explored in~\cite{wang_momentum_2023}.
However, to ensure convergence, the method in~\cite{wang_momentum_2023} requires imposing an additional regularization term to the second moment estimates at every iteration. This is not required in Alada.
The work \cite{crawshaw_robustness_2022} is perhaps the most similar one to ours in the way of reusing first moment estimate. Our theoretical results align with those in \cite{crawshaw_robustness_2022} albeit with a more sophisticated adaptation rule.

\section{Conclusion}
This paper presents Alada, a memory-efficient adaptation method for large-scale matrix optimization.
Alada estimates the second moment of gradients with rank-one factorizations where the factors are adapted alternatively, achieving a sublinear memory cost.
Compared to existing memory-efficient methods such as Adafactor, it involves a first moment estimation rule without introducing memory overheads.
The proposed algorithm is also applicable to tensor-shaped decision variables with only slight modifications. 
We establish the convergence theorem for Alada and also demonstrate its performance on a set of real-world tasks in fine-tuning language models.

\bibliographystyle{IEEEtran}
\bibliography{hxy}

\begin{IEEEbiography}{Xiaoyu He}
	received the B.Eng. degree in computer science and technology from Beijing Electronic Science and Technology Institute, Beijing, China, in 2010, the M.Sc. degree in public administration from South China University of Technology, Guangzhou, China, in 2016, and the Ph.D. degree in computer science from Sun Yat-sen University, Guangzhou, in 2019. He is currently an associate professor in the School of Software Engineering, Sun Yat-sen University. His research interests include optimization and machine learning.
\end{IEEEbiography}

\begin{IEEEbiography}{Yu Cai} completed his undergraduate studies at the School of Computer and Artificial Intelligence, Southwest Jiaotong University, in Chengdu, China, in 2020. He is currently a postgraduate student at the School of Software Engineering, Sun Yat-sen University, in Zhuhai. His primary research focus is machine learning.
\end{IEEEbiography}
\begin{IEEEbiography}{Jin Jia} completed her undergraduate studies at Wuhan University of Science and Technology and pursued a postgraduate degree at Sun Yat-sen University. Her research interests encompass machine learning and agent applications.
	\end{IEEEbiography}
\begin{IEEEbiography}{Canxi Huang} completed his undergraduate studies at the School of Information Science and Technology, Jinan University, in Guangzhou, China, in 2020. He is currently a postgraduate student at the School of Software Engineering, Sun Yat-sen University, in Zhuhai. His primary research focus is machine learning.
	\end{IEEEbiography}

\begin{IEEEbiography}{Wenqing Chen} received the B.E., M.S., and Ph.D. from Huazhong University of Science and Technology, Beijing Normal University, and Shanghai Jiao Tong University in 2012, 2015, and 2022, respectively. He earned his Ph.D. in information and communication engineering. In 2022, he joined Sun Yat-sen University as an assistant professor in the School of Software Engineering. His works have been published in top-tier conferences and journals, including ACL, EMNLP, AAAI, IJCAI, and Expert Syst. Appl. His current research interests lie in the fields of natural language processing and causal inference.
\end{IEEEbiography}

\begin{IEEEbiography}{Zibin Zheng}
	(Fellow, IEEE) is currently a Professor and the Deputy Dean with the School of Software Engineering, Sun Yat-sen University, Guangzhou, China. He authored or coauthored more than 200 international journal and conference papers, including one ESI hot paper and ten ESI highly cited papers. According to Google Scholar, his papers have more than 33,000 citations. His research interests include blockchain, software engineering, and services computing. He was the BlockSys’19 and CollaborateCom16 General Co-Chair, SC2’19, ICIOT18 and IoV14 PC Co-Chair. He is a Fellow of the IET. He was the recipient of several awards, including the Top 50 Influential Papers in Blockchain of 2018, the ACM SIGSOFT Distinguished Paper Award at ICSE2010, the Best Student Paper Award at ICWS2010.
\end{IEEEbiography}

\end{document}